\documentclass[11pt]{article}

\usepackage[utf8]{inputenc}
\usepackage[T1]{fontenc}
\usepackage{lmodern}     
\usepackage{textcomp}
\usepackage{microtype}   
\input{glyphtounicode}   
\usepackage{amsmath}
\usepackage{amssymb}
\usepackage{graphicx}
\usepackage{booktabs}
\usepackage{geometry}
\usepackage[numbers,sort&compress]{natbib}
\usepackage[hidelinks]{hyperref}
\hypersetup{
  pdftitle={AEGIS: A Backup Reflex for Physical AI},
  pdfauthor={Josef Chen},
  pdfsubject={Selective escalation from a frozen-VLA early-warning probe to a stronger separate policy for long-horizon robot manipulation},
  pdfkeywords={physical AI, vision-language-action, robot manipulation, runtime reliability, selective escalation, LIBERO}
}
\usepackage{xcolor}
\usepackage{enumitem}
\usepackage{pifont}   
\usepackage{array}
\usepackage{colortbl}
\usepackage{placeins}   
\newcommand{\cmark}{\textcolor{green!55!black}{\ding{51}}}   
\newcommand{\xmark}{\textcolor{red!70!black}{\ding{55}}}       
\newcommand{\pmark}{\textcolor{orange!85!black}{$\sim$}}        

\newcommand{\RTR}{\mathrm{RTR}}
\newcommand{\Tw}{T_{w}}
\newcommand{\Kmax}{K_{\max}}
\newcommand{\tmin}{t_{\min}}
\newcommand{\Rmax}{R_{\max}}
\newcommand{\hh}{\mathbf{h}}            
\newcommand{\xx}{\mathbf{x}}            
\newcommand{\sscore}{s}                 
\newcommand{\Esc}{\mathsf{esc}}         
\newcommand{\Real}{\mathbb{R}}
\newcommand{\indic}{\mathbf{1}}
\DeclareMathOperator*{\meanpool}{mean\text{-}pool}
\DeclareMathOperator*{\MAD}{MAD}
\DeclareMathOperator*{\Var}{Var}

\title{\textbf{AEGIS}: A Backup Reflex for Physical AI\\[2pt]
\large Calling a Stronger Policy Before Long-Horizon Failures Compound}

\author{%
  Josef Chen \\
  KAIKAKU \\
  \texttt{josef@kaikaku.ai}
}
\date{June 2026}

\begin{document}
\maketitle

\begin{abstract}
Long-horizon robot manipulation tends to fail gradually: one bad step degrades the
state, and the policy spirals into a basin from which it cannot recover. The failure
is visible before it happens. A cheap probe on the policy's own frozen activations
predicts it early, while there is still time to act. We introduce \textbf{AEGIS}
(Activation-probe Early-warning, Gated Inference Switching): when the probe flags a
step, control switches to a stronger separate policy, but only for the steps that
need it. The thesis is one sentence. A robot policy can read its own activations as
an early-warning signal and call a stronger policy before failure compounds,
recovering twice as many failures as matched-budget escalation.

On LIBERO-Spatial, AEGIS recovers $10.1\%$ of the trajectories the weak policy alone
loses, against $4.6\%$ for budget-matched blind escalation and $5.1\%$ for a
random-trigger placebo (one-sided exact paired McNemar tests, Holm--Bonferroni
adjusted over the three pre-registered contrasts: $+5.4$pp over blind,
$p=8.5{\times}10^{-6}$; $+5.0$pp over random, $p=1.0{\times}10^{-4}$; paired-trajectory
bootstrap CIs exclude zero). It does this while the stronger policy is active on only
$38\%$ of steps (its duty cycle), so the lever is timing, not compute: the same
selectivity that recovers tasks is what lets the stronger $4.14$B policy stay dormant
most of the time. The probe clears its precondition with an early-window AUROC of
$0.764$ ($95\%$ CI $[0.70, 0.84]$), read from the weak-policy path over the first
$30\%$ of trajectory steps before any handoff. We
pre-register the full analysis plan, including a conditional recovered-task-rate
estimand and explicit kill criteria, and confirm the result on $700$
common-random-number episodes per arm ($n_{A\text{-fail}}{=}646$).
\end{abstract}

\section{Introduction}
\label{sec:intro}

\begin{figure}[t]
  \centering
  \includegraphics[width=0.78\linewidth]{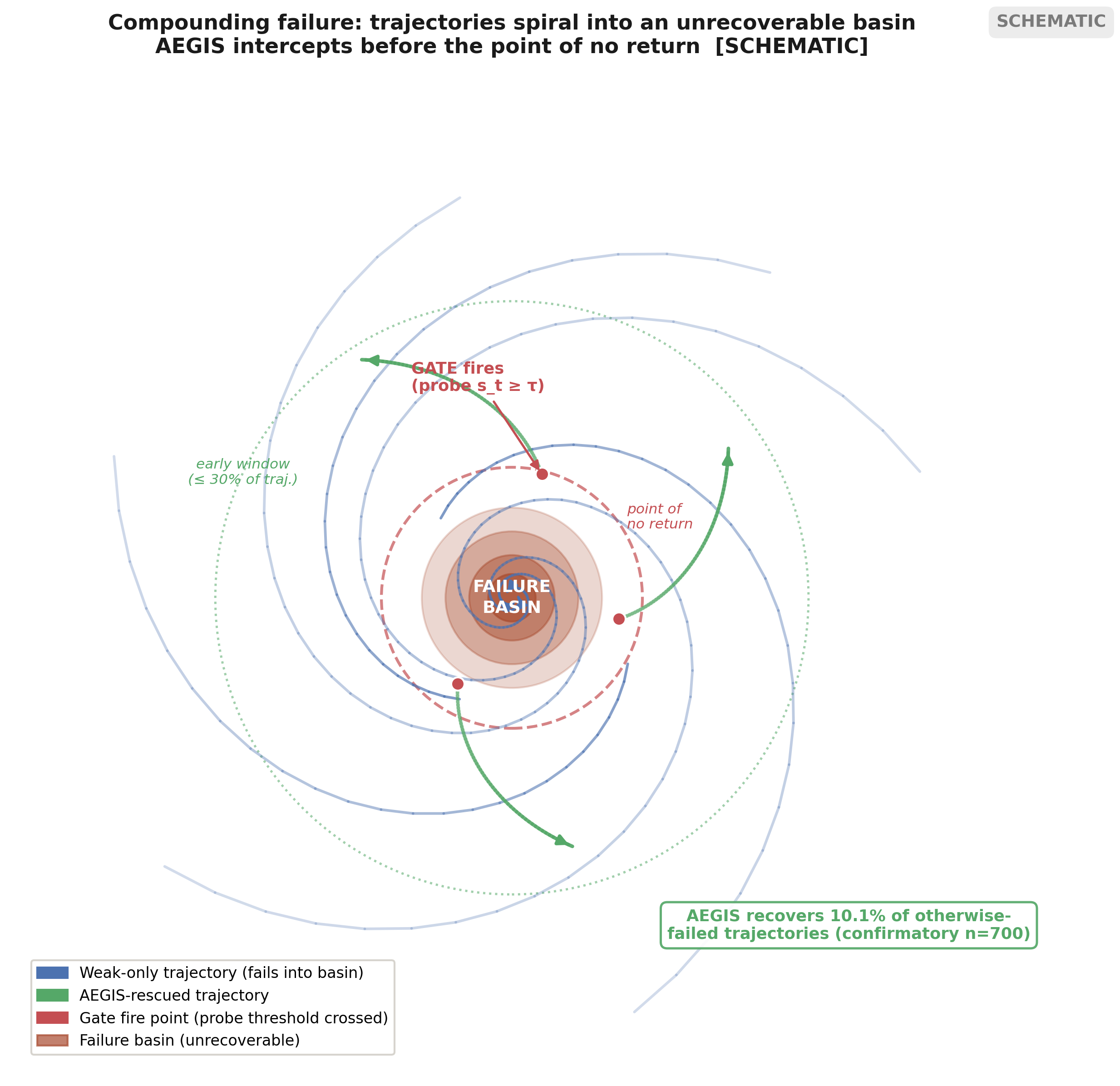}
  \caption{\textbf{Why timing is the whole problem.} Schematic phase portrait of
  long-horizon manipulation: under the weak policy alone, a perturbed trajectory spirals
  inward and compounds toward an unrecoverable failure basin (shaded). Recovery is only
  possible while the trajectory is still outside the point-of-no-return ring (dashed). The
  AEGIS probe fires the gate within the early window ($\leq 30\%$ of trajectory steps,
  red points) and hands control to the stronger policy, deflecting otherwise-doomed
  trajectories back outward (green), recovering $10.1\%$ of episodes the weak policy alone
  fails (confirmatory $n{=}700$).
  \emph{(Schematic phase portrait; recovery rate is a measured value, Section~\ref{sec:results}.)}}
  \label{fig:spiral}
\end{figure}

A robot policy rarely fails all at once. A long-horizon manipulation failure is a
slow spiral (Fig.~\ref{fig:spiral}): one mistimed grasp nudges the arm
off-distribution, the next action compounds the error, and within a few steps the
trajectory has crossed a point of no return. The warning arrives long before the
crash. The policy's own activations betray the coming failure while there is still
time to act. We call the missing response a \emph{runtime authorization gap}: the
absence of a layer that decides, at run time, what a policy is and is not allowed to
do as evidence of impending failure mounts. Two research programs now populate this gap.
Between them they leave one axis conspicuously empty.

The first program is \emph{detect-only} failure prediction: read a cheap signal from
the policy and raise an alarm. SAFE trains a probe on hidden states and predicts
manipulation failure at AUROC $72$--$93\%$ on LIBERO, then halts or calls a human
\citep{safe2025}; FIPER, FAIL-Detect, ReconVLA, and Sentinel raise conformal or
consistency-based alarms in the same spirit \citep{fiper2025, faildetect2025,
reconvla2026, sentinel2024}. These methods own \emph{prediction}; none of them
\emph{act} to recover, and none report a recovery metric. The second program is
\emph{recover-within-the-same-policy}: when an alarm fires, do something with the
policy you already have. HELM augments and recovers within the same frozen policy
\citep{helm2026}; Pre-VLA resamples the same policy behind a
warm-up gate, recovering $+6.83$pp \citep{prevla2026}; LiLo-VLA retries and backtracks
\citep{lilovla2026}; FailSafe and FPC-VLA author corrective actions or strategies for
the same motor stack \citep{failsafe2025, fpcvla2025}. These methods own
\emph{recovery}, but they resample, replan, or re-prompt the \emph{same} policy that
is failing.

Notice what a human supervisor would do. Detect-only methods see the spiral but
cannot act. Recover-in-policy methods act, but only by asking the same failing
policy to try again. Neither calls for help. AEGIS does. When the cheap policy is
about to fail, call in a stronger one, but only for the steps that need it. We call
this mechanism \textbf{AEGIS} (Activation-probe Early-warning, Gated Inference
Switching): a frozen-VLA per-step early-warning probe whose flagged steps are
escalated, with control switched mid-trajectory, to a stronger \emph{separate}
policy, and the resulting recovery measured directly and defended with causal
controls. AEGIS is the outward counterpart to our companion memory gate AURA-Mem
\citep{auramem2026}, which gates writes \emph{inward} at fixed compute; AEGIS
switches in a stronger policy \emph{outward}, and only on the steps a cheap signal
flags. To our knowledge, prior work has not yet evaluated this exact combination: early prediction from frozen internals,
threshold-triggered escalation to a stronger separate policy, a measured recovery
metric, and causal controls that the gain is selectivity rather than spent compute.
We lay this out against the field in Table~\ref{tab:positioning}.

Accurate prediction does not imply effective prevention. In the language-model
setting, a probe at AUROC $0.94$ can still \emph{reduce} task success by $26$pp when
it triggers interventions that disrupt trajectories which would otherwise have
succeeded \citep{interventionparadox2026}, a cautionary result we import from the LLM
domain rather than measure on our own policies. So a method that escalates has to
prove its gains come from \emph{where} it escalates, not from the extra compute it
spends. A budget-matched blind-escalation control and a random-trigger placebo
isolate exactly that.

We use \emph{physical AI} in the narrow sense of embodied policies that map perception and
language-conditioned task context to robot actions. The timing matters now because of how
such policies will be deployed. As they scale,
deployment will look less like choosing one policy and more like scheduling a
hierarchy of policies under latency and compute constraints. A cheap policy drives
most of the time; a frontier policy is too expensive to run constantly, because
single-stream robot decode is memory-dominated and its cost is paid per active policy
\citep{chen2026memorybound}, and so gets called selectively. The central question becomes: when does the cheap policy still
deserve control, and when should a stronger one take over? AEGIS is a concrete answer
to that question, and the controls below are what turn the answer into evidence.

\begin{figure}[t]
  \centering
  \includegraphics[width=0.86\linewidth]{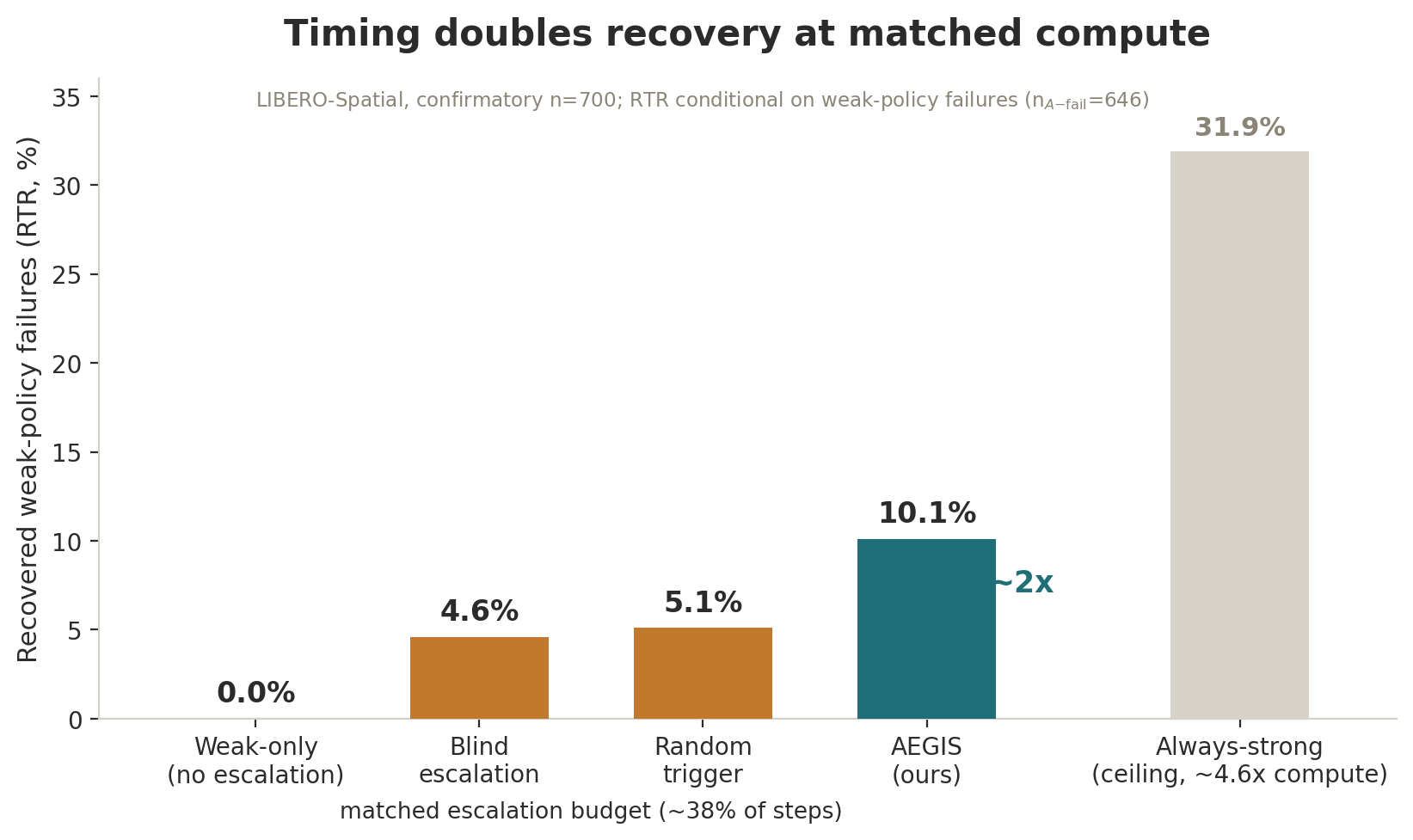}
  \caption{\textbf{Timing doubles recovery at matched compute.} Conditional
  recovered-task rate (RTR) on the weak-policy-failing subset of LIBERO-Spatial
  (confirmatory $n{=}700$; $n_{A\text{-fail}}{=}646$). At a shared escalation budget
  (about $38\%$ of steps), AEGIS recovers $10.1\%$ of otherwise-failed trajectories,
  roughly twice the budget-matched blind ($4.6\%$) and random-trigger ($5.1\%$)
  controls. The always-strong arm (grey) marks the recovery ceiling but pays roughly
  $4.6\times$ the compute. Selectivity, not spend, is the active ingredient.}
  \label{fig:headline}
\end{figure}

\paragraph{Contributions.}
\begin{itemize}[leftmargin=1.4em, itemsep=2pt, topsep=2pt]
  \item \textbf{A new runtime escalation problem.} When should a cheap robot policy
        call a stronger policy, before failure compounds rather than after it
        completes (\S\ref{sec:intro}, \S\ref{sec:related})?
  \item \textbf{A concrete mechanism.} A frozen hidden-state probe gates step-level
        escalation to a stronger \emph{separate} policy, with an early-harm gate, a
        conformal trigger threshold, and a per-episode budget cap (\S\ref{sec:method}).
  \item \textbf{A causal test of selectivity.} Budget-matched blind escalation and a
        random-trigger control show the gain comes from \emph{where} escalation
        happens, not from extra compute (\S\ref{sec:design}, \S\ref{sec:results}).
  \item \textbf{A measured deployment tradeoff.} AEGIS recovers $10.1\%$ of
        weak-policy failures while escalating on $38\%$ of steps, roughly doubling
        matched-budget recovery (Fig.~\ref{fig:headline}, \S\ref{sec:results}).
\end{itemize}

\section{Related Work}
\label{sec:related}

We organize prior art along the two programs that flank the gap this paper fills:
methods that \emph{detect} impending failure but do not act on it, and methods that
recover but only by resampling, replanning, or re-prompting the same policy. To our
knowledge, prior work has not yet evaluated the combination of per-step early failure
prediction from frozen VLA internals, escalation to a stronger and separate policy, a
measured recovery metric, and causal controls. Detect-only owns axis (1); recover-within-policy owns axis (3)
but resamples or replans the same policy. The axis of escalating flagged steps to a
stronger separate policy is unoccupied. Table~\ref{tab:positioning} makes the gap
visual: AEGIS is the only entry that combines all four capabilities at once.

\begin{table}[htbp]
  \centering
  \small
  \setlength{\tabcolsep}{5pt}
  \renewcommand{\arraystretch}{1.25}
  \caption{Where AEGIS sits relative to representative prior art. Columns are the four
  capabilities that, taken together, define the unoccupied axis. \cmark~= yes,
  \xmark~= no, \pmark~= partial. AEGIS is the only row with all four. ``Stronger
  \emph{separate} policy'' means the intervention runs a more capable
  distinct policy, not a resample, replan, or re-prompt of the failing one. Each mark
  reflects only what the cited paper reports: SAFE detects (AUROC $72$--$93\%$ on LIBERO)
  and halts or defers (\S\ref{sec:related-detect}); Pre-VLA resamples the same policy for
  $+6.83$pp (\S\ref{sec:related-recover}); HELM rolls back and replans within the same
  frozen policy (\S\ref{sec:related-recover}); FailSafe conditions recovery actions and
  reports $+22.6\%$ on ManiSkill (\S\ref{sec:related-recover}). The escalation column is
  marked \cmark{} only when the recovery routes to a distinct stronger executor, which is
  why no prior row earns it.}
  \label{tab:positioning}
  \begin{tabular}{l c c c c}
    \toprule
     & \textbf{Early} & \textbf{Escalates to a} & \textbf{Measures} & \textbf{Causal} \\
    \textbf{Method} & \textbf{($\leq\!30\%$)} & \textbf{stronger} & \textbf{task} & \textbf{controls} \\
     & \textbf{frozen probe} & \textbf{\emph{separate} policy} & \textbf{recovery} & \textbf{(selectivity)} \\
    \midrule
    SAFE \citep{safe2025}               & \cmark & \xmark & \xmark & \xmark \\
    FIPER / FAIL-Detect \citep{fiper2025,faildetect2025} & \pmark & \xmark & \xmark & \xmark \\
    Sentinel \citep{sentinel2024}       & \pmark & \xmark & \xmark & \xmark \\
    INSIGHT (detect+defer) \citep{insight2025} & \pmark & \xmark & \xmark & \xmark \\
    \midrule
    HELM \citep{helm2026}               & \xmark & \xmark & \cmark & \pmark \\
    Pre-VLA \citep{prevla2026}          & \pmark & \xmark & \cmark & \xmark \\
    LiLo-VLA \citep{lilovla2026}        & \xmark & \xmark & \cmark & \xmark \\
    FailSafe / FPC-VLA \citep{failsafe2025,fpcvla2025} & \xmark & \xmark & \cmark & \xmark \\
    Always-strong (compute ceiling)     & -- & \cmark & \cmark & \xmark \\
    \midrule
    \rowcolor{green!8}
    \textbf{AEGIS (ours)}               & \cmark & \cmark & \cmark & \cmark \\
    \bottomrule
  \end{tabular}
\end{table}

\subsection{Detect-only failure prediction}
\label{sec:related-detect}

Our signal's closest twin is SAFE, which trains a probe on a manipulation policy's
hidden states and predicts eventual failure at AUROC $72$--$93\%$ on LIBERO
\citep{safe2025}. We adopt the same family of cheap internal signal, a hidden-state
probe read from a frozen policy, and we report the same precondition quantity
(early-window AUROC). The difference is what happens after the alarm: SAFE halts or
defers to a human, and reports no recovery. We treat SAFE's all-steps AUROC as the
reference for our early-window ($\leq 30\%$) precondition and then go past it,
escalating and measuring recovery. Several methods raise a conformal alarm from
different sources. FIPER uses RND-based out-of-distribution scores and action entropy
\citep{fiper2025}. FAIL-Detect uses a success-only OOD detector with flow-density
scoring \citep{faildetect2025}. ReconVLA places its alarm on action tokens
\citep{reconvla2026}. Sentinel fuses temporal action-consistency with a
vision-language model for early warning \citep{sentinel2024}. Pre-VLA additionally
exposes a validity head used for detection \citep{prevla2026}. Every method in this
group \emph{detects}: it predicts or flags, and then halts, defers, or resamples.
INSIGHT is a close detect-and-defer neighbor: token-level uncertainty triggers a
request for help rather than a separate stronger executor \citep{insight2025}. None
escalates to a stronger separate policy, and none reports a recovered-task-rate. Casting
this absence as a runtime-authorization gap raises the bar on what a method must
demonstrate to claim it closes the gap rather than merely measuring it.

\subsection{Recover-within-the-same-policy}
\label{sec:related-recover}

The closest prior work overall is HELM, which recovers within the \emph{same} frozen
policy instead of escalating to a different one
\citep{helm2026}. HELM is both our nearest neighbor and a required foil: we re-implement
a rollback-to-checkpoint-and-replan recovery in the spirit of recover-within-the-same-policy
methods as a budget-matched baseline arm
(\S\ref{sec:method}) and require targeted escalation to beat it at matched compute. The
distinction is the policy class. HELM never leaves the policy that is failing; we
route to a stronger one. Pre-VLA resamples the same policy behind a
real warm-up horizon $\Tw$ before it allows intervention, recovering $+6.83$pp
\citep{prevla2026}. We adopt
the idea of a warm-up horizon as one motivation for our early-harm gate while
differentiating its mechanism: we gate escalation, not verification activation,
and add a per-episode budget cap. LiLo-VLA recovers on LIBERO-Long by retrying and
backtracking within a modular planner-plus-VLA stack \citep{lilovla2026}. On a different
benchmark, FailSafe conditions recovery actions on LLaVA-OV and reports $+22.6\%$ on
ManiSkill, again as recovery actions inside the policy rather than escalation to a
separate one \citep{failsafe2025}. A VLM supervisor in FPC-VLA \emph{authors corrective
strategies} for the same motor stack, not a stronger motor policy
\citep{fpcvla2025}. Confidence-Gated Robot Autonomy gates between acting and deferring
on uncertainty \citep{confidencegated2026}; deferring is not escalating, since there is no
stronger executor that takes over. Finally, FARL learns recovery via an RL post-training
world-model safety critic with offline recovery, a training-time regime distinct from
our runtime, training-free escalation \citep{farl2026}. World-model approaches learn to
imagine or roll out future states for anticipation or replanning, whether as a unified
VLA-plus-world-model \citep{rynnvla2025} or an RL post-training safety critic
\citep{farl2026}. Our early-warning signal instead reads directly from \emph{frozen}
VLA internals, with no learned dynamics model and no training-time regime. And our
intervention escalates to a stronger separate policy rather than replanning against an
imagined rollout, the recover-within-the-same-policy lane we use as a baseline through
HELM and Pre-VLA \citep{helm2026, prevla2026}.

\subsection{Relation to the authors' companion memory work, and theoretical anchors}
\label{sec:related-aura}

This paper is the orthogonal inverse of our companion memory work, AURA-Mem
\citep{auramem2026}. AURA gates memory writes \emph{inward} to save bandwidth at fixed
success; AEGIS gates compute and policy \emph{outward} to raise success at
fixed memory. The trigger semantics differ (should I write versus will
this trajectory fail and should I escalate) and so does the metric. AURA itself notes
that it is a memory layer that does not by itself raise robot success, and this
paper is the success-raising counterpart. Any action-prediction complement we
compute is retrained on the trajectory-failure label, never on AURA's write-worthiness
target.

Our three causal controls and early-harm gate are motivated by the
intervention-paradox result: accurate prediction does not imply
effective prevention, and an AUROC-$0.94$ predictor can drive a $-26$pp change in
success because interventions that recover failing trajectories also disrupt
trajectories that would have succeeded \citep{interventionparadox2026}. This is the
motivation, not a claim we make about our own numbers. The compounding nature of the
failures we target is the classical $O(\epsilon T^2)$ error-accumulation intuition of
DAgger \citep{dagger2011}. Our substrate is LIBERO \citep{libero2023}; our probe-target
backbone and the supporting quantization experiment use OpenVLA-OFT
\citep{openvlaoft2025}; RynnVLA-002 is cited for context as a unified VLA-plus-world-model
backbone \citep{rynnvla2025}.

\section{Method}
\label{sec:method}

\begin{figure}[htbp]
  \centering
  \includegraphics[width=\linewidth]{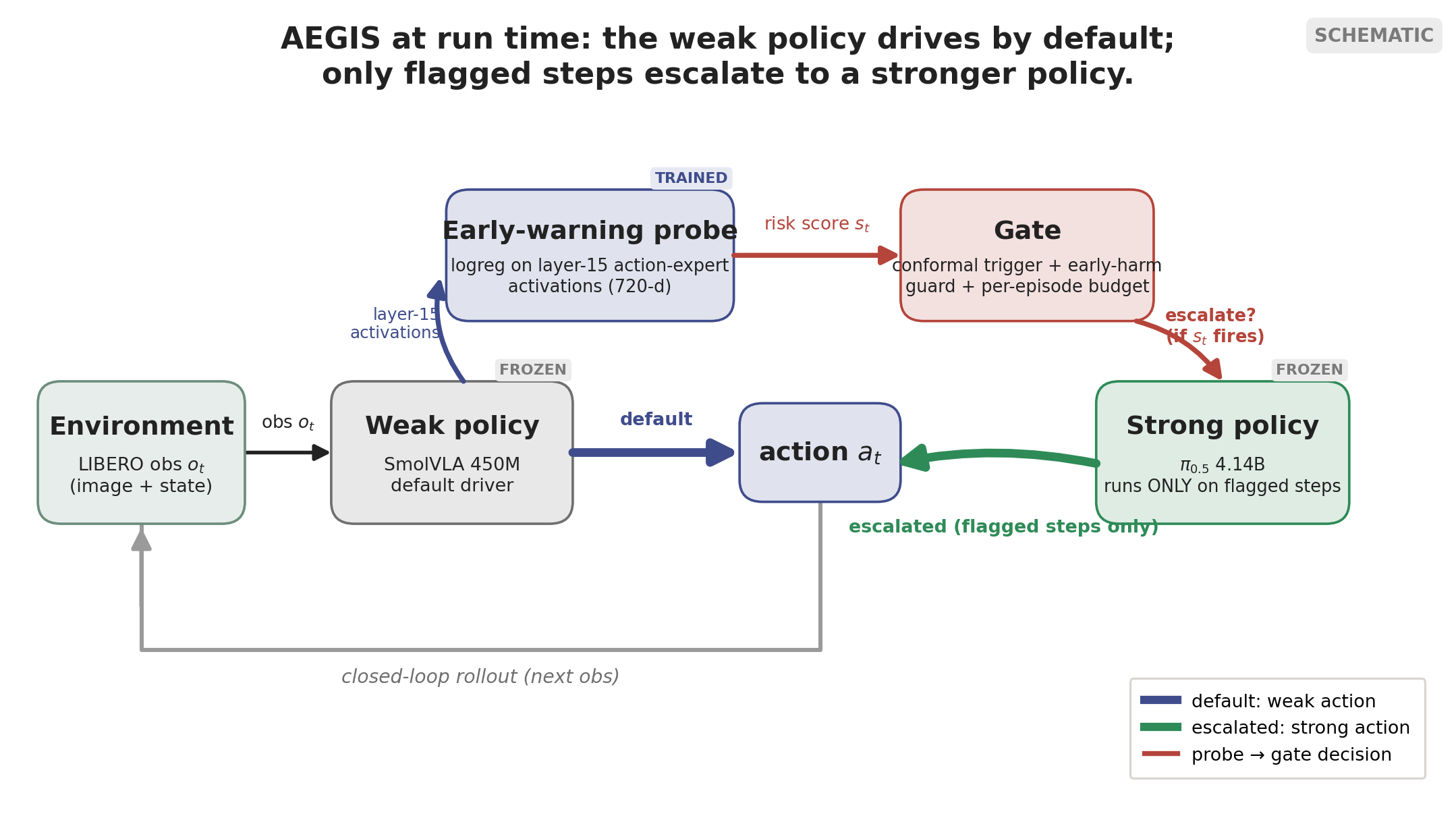}
  \caption{\textbf{AEGIS at run time.} The weak policy drives by default (blue path).
  A logistic-regression probe (the only trained component) reads the weak policy's
  frozen layer-15 action-expert activations and emits a per-step risk score. A gate
  (conformal trigger $+$ early-harm guard $+$ per-episode budget cap) turns that score
  into a binary escalation decision; on flagged steps, control switches at the next chunk
  boundary to a stronger \emph{separate} frozen policy (green path), which
  runs only while flagged. The emitted action $a_t$ steps the environment and closes the
  loop back to the next observation (bottom arrow). Both policies stay frozen; nothing is
  fine-tuned.}
  \label{fig:architecture}
\end{figure}

\begin{figure}[t]
  \centering
  \includegraphics[width=0.95\linewidth]{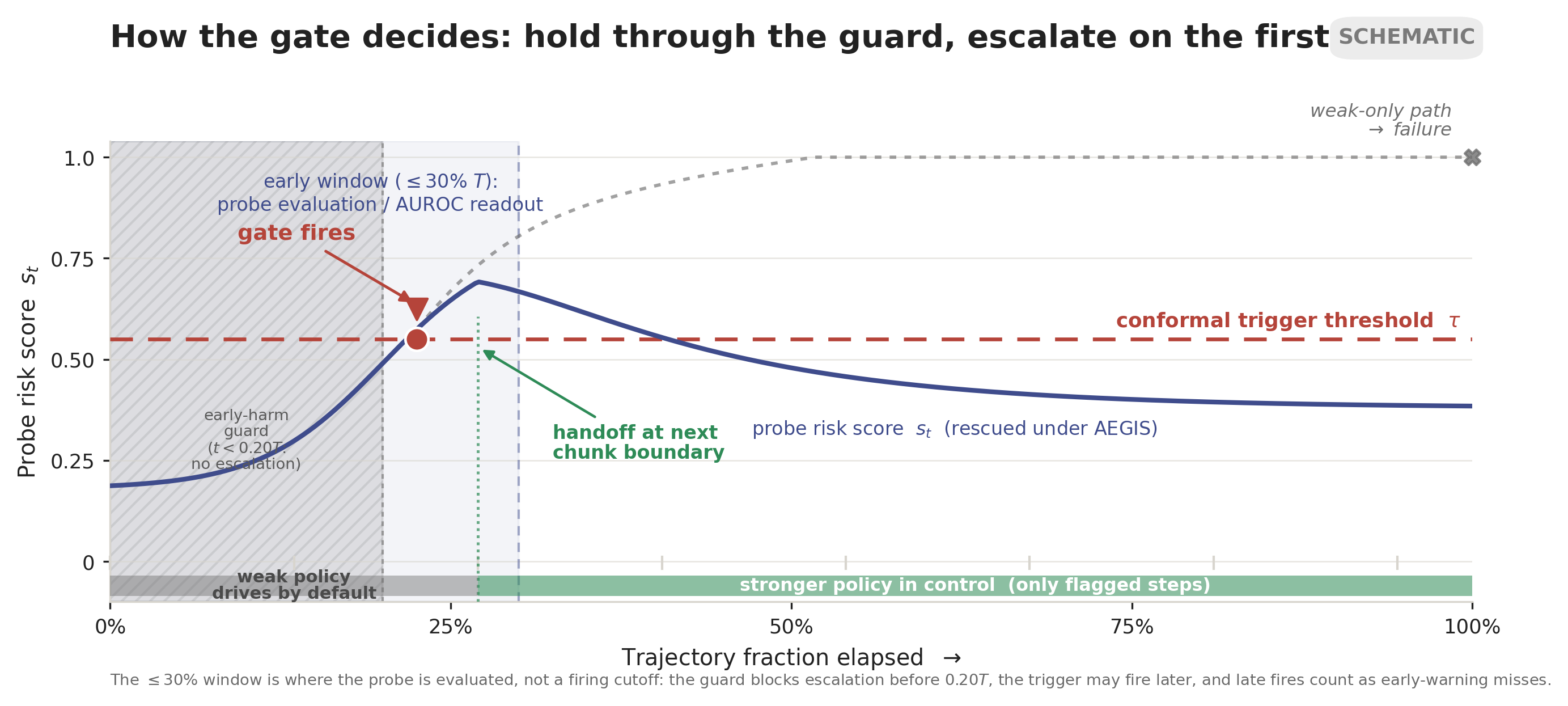}
  \caption{\textbf{How the gate decides.} Stylized per-step view of one trajectory. The
  probe's risk score $\sscore_t$ (indigo) rises as failure approaches. The early-harm guard
  suppresses any escalation before $\tmin{=}0.20\,T$ (hatched); the $\leq 30\%$ band is the
  probe's evaluation window where AUROC is read, not a runtime firing cutoff. When
  $\sscore_t$ crosses the conformal threshold $\tau$ (red), control hands to the stronger
  policy at the next chunk boundary (green) and is held, deflecting the trajectory that the
  weak policy alone would have driven to failure (dashed grey). The trigger may also fire
  later than this example; late fires still act but count as early-warning misses. The same
  selectivity is what keeps the stronger policy dormant on the other steps.}
  \label{fig:timeline}
\end{figure}

AEGIS has four parts (Fig.~\ref{fig:architecture}): a cheap per-step signal read from a frozen
VLA (\S\ref{sec:method-signal}), a conformal threshold that turns the signal into a
trigger at a controlled false-trigger rate (\S\ref{sec:method-threshold}), an early-harm
gate that suppresses premature and excessive escalation
(\S\ref{sec:method-threshold}), and an escalation handoff that routes flagged steps to a
stronger separate policy (\S\ref{sec:method-escalation}). Figure~\ref{fig:timeline} shows
the gate logic on a single trajectory: the risk score rises, crosses the conformal
threshold inside the early window, and hands control to the stronger policy at the next
chunk boundary. \S\ref{sec:method-causal} then
gives the causal-identification argument that makes the resulting recovery a claim about
the \emph{selectivity} of the signal rather than about spending extra compute.

\paragraph{Setup and notation.}
A deployed (\emph{weak}) policy $\pi_w$ executes a long-horizon task as a trajectory of
$T$ control steps; it emits actions in chunks of horizon $H$, so step $t$ belongs to
chunk $c(t) = \lfloor t/H \rfloor$. The pre-registration described $H = 50$; the executed
checkpoints integrate $10$-step action chunks, so the reported runs use $H = 10$ to match
the policies' native granularity (this deviation is logged in \S\ref{sec:deviations} and
changes no estimand or contrast definition). A stronger \emph{separate} policy
$\pi_s$ is available but is not run by default. At each step $t$ the controller observes a
cheap per-step score $\sscore_t \in \Real$ read from $\pi_w$'s frozen internals
(\S\ref{sec:method-signal}), compares it to a calibrated, time-varying threshold
$\delta_t$ (\S\ref{sec:method-threshold}), and emits a binary escalation decision
$\Esc_t \in \{0,1\}$ subject to an early-harm gate and a per-episode budget cap. When
$\Esc_t{=}1$ the next chunk boundary hands control to $\pi_s$
(\S\ref{sec:method-escalation}). Both $\pi_w$ and $\pi_s$ remain \emph{frozen}: no
policy weights are ever updated, and the only learned component is the probe head, whose
parameters are disjoint from both policies.

\subsection{Cascade signal from frozen VLA internals}
\label{sec:method-signal}

\paragraph{Primary signal: a hidden-state failure probe.}
The primary signal is a probe read from $\pi_w$'s frozen internal activations. We place a
forward hook on a fixed layer of the deployed policy and mean-pool the captured
activations over the $H$-token action chunk to obtain a fixed-dimensional feature
\[
  \hh_t \;=\; \meanpool_{j=1}^{H}\, a^{(\ell)}_{t,j} \;\in\; \Real^{d},
\]
where $a^{(\ell)}_{t,j}$ is the layer-$\ell$ activation at token $j$ of the chunk at step
$t$. The deployed policy is SmolVLA (a $450$M VLA) \citep{smolvla2025}. The probe reads the activations of the
\emph{action expert} (the head that integrates flow-matching action chunks), not the
vision encoder: we hook the output projection of the self-attention block at action-expert
layer $\ell = 15$, i.e.\ \texttt{model.\allowbreak vlm\_with\_expert.\allowbreak lm\_expert.\allowbreak layers[15].\allowbreak self\_attn.\allowbreak o\_proj},
which exposes a $720$-dimensional per-token representation; mean-pooling over the chunk
gives $\hh_t \in \Real^{720}$ ($d = 720$). This hook is read \emph{live} during rollout
(the captured activations vary step-to-step, standard deviation $> 0.05$, confirming the
hook fires on the live forward pass rather than on a frozen cached feature). An earlier
implementation that hooked the vision encoder captured a feature that did not vary with the
rollout and yielded chance-level prediction (AUROC $0.50$); we identified this as a
hook-placement bug, moved the probe onto the live action-expert path above, and report the
bug and its correction in \S\ref{sec:deviations}. The probe head is a two-layer multilayer
perceptron $g_\theta:\Real^{d}\!\to\!(0,1)$ \citep{alainbengio2016} with architecture $[d \to 256 \to 1]$ (i.e.\
$[720\to256\to1]$ on the executed SmolVLA action-expert path; $[4096\to256\to1]$ when
probing an OFT-7B backbone in the supporting study \citep{openvlaoft2025}), trained with a
binary cross-entropy
objective and per-class weights against the \emph{eventual} trajectory outcome label
$y \in \{0,1\}$ (fail $=1$). The per-step probe score is
$\sscore^{\mathrm{probe}}_t = g_\theta(\hh_t)$.

\paragraph{Label, early window, and the trajectory-split protocol.}
The label $y$ is the trajectory's eventual success/failure, so the probe is trained to
\emph{anticipate} failure rather than to react to it. Two leakage-control choices are made
before any data is seen and are binding. First, the probe is trained \emph{only} on early
steps, $t \leq 0.30\,T$, so that prediction by the readout point cannot use information
from the part of the trajectory the controller is trying to pre-empt; AUROC is reported on
this same early window. Second, the train/validation/test split is performed at the
\emph{trajectory} level, $70/15/15$, never at the step level: all early-window steps of a
given rollout fall entirely within one split, which prevents the temporally correlated
steps of one trajectory from appearing in both training and test. All probe fitting and
AUROC scoring are done offline on logged rollouts, so the signal adds no rollout-time cost
beyond the forward hook. The precondition target is early-window
$\mathrm{AUROC}(0.30\,T) \geq 0.75$ with a DeLong confidence interval; SAFE's all-steps
AUROC of $72$--$93\%$ is the reference, and we report the \emph{early-window} curve rather
than an all-steps number \citep{safe2025}. That a linear or shallow probe on frozen VLA
internals carries outcome-relevant information is what we rely on here. Our contribution is
its use as an \emph{early}, conformally-thresholded \emph{escalation} trigger
with a recovered-task-rate readout and causal controls.

\begin{figure}[htbp]
  \centering
  \includegraphics[width=\linewidth]{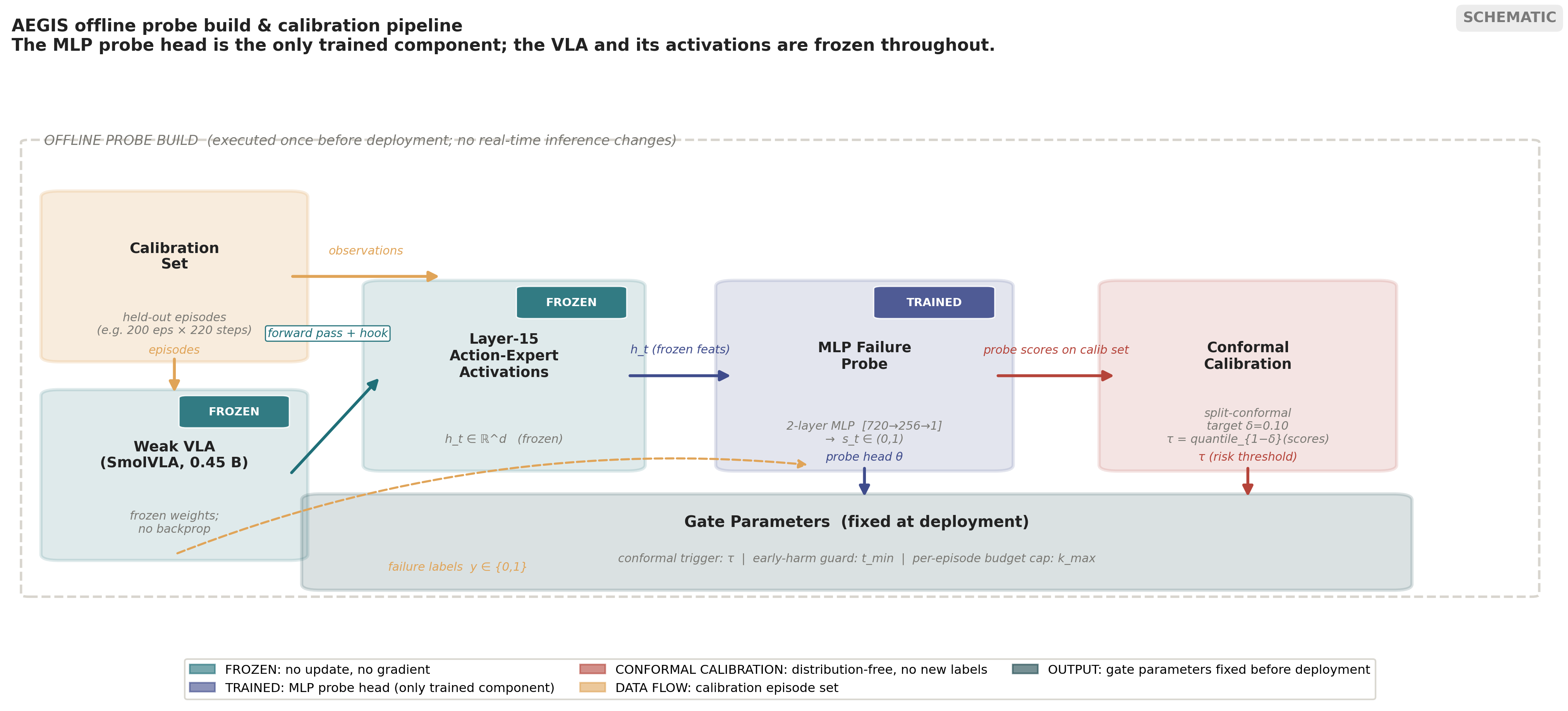}
  \caption{\textbf{Offline probe build and calibration.} Everything in this figure happens
  \emph{once, before deployment}, and changes nothing at rollout time. A held-out calibration
  set is rolled out under the frozen weak VLA; a forward hook captures layer-15 action-expert
  activations $\hh_t$. The only trained component is a two-layer MLP probe head
  $[720\to256\to1]$ fit to the eventual-failure label on early-window steps. Split-conformal
  calibration then turns the probe scores into a distribution-free risk threshold $\delta_t$
  (target miscoverage), and the resulting gate parameters (conformal trigger, early-harm
  guard $t_{\min}$, per-episode budget cap $k_{\max}$) are frozen for deployment. Both
  policies stay frozen throughout. \emph{(Schematic of the offline pipeline; numerical
  hyperparameters are given in \S\ref{sec:method-threshold}.)}}
  \label{fig:probe-pipeline}
\end{figure}

\paragraph{Cheap complementary signals.}
Alongside the probe we compute three complementary scores from the same single logged
rollout, all essentially free or cheap:
\begin{itemize}[leftmargin=1.4em, itemsep=2pt, topsep=2pt]
  \item \textbf{Chunk-overlap delta} $\Delta^{\mathrm{chunk}}_t$: the $L_2$ disagreement
        between the overlapping regions of consecutive action chunks,
        $\Delta^{\mathrm{chunk}}_t = \lVert a_{t} - a_{t-1}\rVert_2$ over the overlap.
        Literally free from the logged actions.
  \item \textbf{STAC variance} $\Var^{\mathrm{STAC}}_t$: the temporal-consistency variance
        of the action chunk. Free.
  \item \textbf{Flow-matching sampling disagreement} $\mathrm{MAD}^{K}_t$: we draw
        $K = 4$ action chunks from $\pi_w$ with distinct flow-matching noise seeds and take
        the mean absolute deviation across samples,
        $\mathrm{MAD}^{K}_t = \MAD\big(\{a^{(k)}_t\}_{k=1}^{K}\big)$. On SmolVLA/$\pi_{0.5}$
        this reuses a shared KV cache so only the small action expert re-integrates; the
        added wall-clock is reported as measured rather than pre-claimed.
\end{itemize}
The fused per-step feature is $\xx_t = [\,\sscore^{\mathrm{probe}}_t,\,
\Delta^{\mathrm{chunk}}_t,\, \Var^{\mathrm{STAC}}_t,\, \mathrm{MAD}^{K}_t\,]$, and the
trigger score $\sscore_t$ used downstream is the probe output by default; fusion of the
complements beyond the primary probe is exploratory and labelled as such. We explicitly do
\emph{not} rely on token-logit entropy or MC-dropout as primary signals (both sit near
chance on these policies), and we do not use a quantile-spread head, which on the
$\pi_{0.5}$ checkpoint is a normalization artifact rather than an uncertainty estimate.

\paragraph{Differentiation experiment.}
On the \emph{same} rollouts we also compute an action-prediction-surprise signal in the
style of our companion memory work \citep{auramem2026}, retrained on the
trajectory-failure label, never on the write-worthiness target. The closest surprise proxy
we log, the chunk-overlap delta $\Delta^{\mathrm{chunk}}$, carries some early signal
(early-window AUROC $0.63$) but is materially weaker than the failure-anticipatory probe
($0.738$ conservative pilot, $0.764$ on the confirmatory run). The surprise proxy is
informative but dominated. The differentiation from the companion memory work therefore does not
rest on surprise being at chance. It rests on \emph{orthogonal trigger semantics}
(should I write to memory versus will this trajectory fail and should I
escalate) and an orthogonal metric (write-worthiness versus recovered-task-rate), with
the failure-trained probe additionally being the stronger early predictor. This is the
empirical reason the probe is more than a relabelled action-prediction signal.

\subsection{Conformal trigger threshold and early-harm gate}
\label{sec:method-threshold}

\paragraph{Split-conformal trigger threshold.}
The full offline build that produces this threshold and the gate parameters is summarized in
Fig.~\ref{fig:probe-pipeline}. We convert the per-step score $\sscore_t$ into a trigger with split-conformal calibration \citep{conformalintro2021}.
Using a held-out calibration set of early-window steps drawn from would-not-fail
trajectories, we treat $\sscore_t$ as a nonconformity score and choose a time-varying
threshold $\delta_t$ as the appropriate empirical quantile of the calibration scores so
that the per-step false-trigger rate is controlled at level $\alpha = 0.10$; concretely,
with $n$ calibration scores the threshold is the $\lceil (1-\alpha)(n+1)\rceil$-th order
statistic. We use this as split-conformal-\emph{style} quantile calibration to target a
nominal per-step false-trigger rate, and we are deliberately careful about what it
guarantees: because steps within a trajectory are correlated and the threshold $\delta_t$
varies with trajectory fraction, the finite-sample coverage is marginal at the calibrated
score level rather than a clean trajectory-level distribution-free guarantee, so we report
the realized trigger rates empirically rather than leaning on the nominal level. The raw
escalation indicator is then
\[
  \Esc^{\mathrm{raw}}_t \;=\; \indic\!\left[\,\sscore_t \,\ge\, \delta_t\,\right].
\]
We allow $\delta_t$ to vary with trajectory fraction $t/T$ so that the false-trigger rate
is controlled across the early window rather than only on average. We sweep
$\alpha \in [0.01, 0.20]$ to trace the overhead/recovery tradeoff curve (descriptive).

\paragraph{Per-stratum calibration.}
Marginal conformal coverage need not hold \emph{conditionally} within a difficulty
stratum. We therefore calibrate one threshold per difficulty stratum (the terciles of
\S\ref{sec:design}) whenever the per-stratum calibration set is large enough; where a
stratum is too small to calibrate its own threshold without instability, we fall back to a
shared threshold and explicitly disclaim that coverage is then marginal rather than
conditional. The choice between per-stratum and shared calibration is decided on the pilot
and reported, not chosen after seeing main-run outcomes.

\paragraph{Early-harm gate and budget cap.}
An \emph{early-harm gate} suppresses any escalation before $\tmin = \max(0.20\,T, 2)$,
and a per-episode budget cap admits at most $\Kmax = \lceil 0.05\,T \rceil$ escalations,
ranked by signal within the budget. The realized escalation decision is therefore
\[
  \Esc_t \;=\; \Esc^{\mathrm{raw}}_t \cdot \indic[\,t \ge \tmin\,] \cdot
           \indic\!\Big[\textstyle\sum_{u \le t}\Esc_u \le \Kmax\Big],
\]
with ties at the budget boundary broken by the larger $\sscore_u$. The gate is motivated
both by the intervention-paradox finding \citep{interventionparadox2026} and by Pre-VLA's
real warm-up horizon $\Tw$
\citep{prevla2026}. We differentiate our gate from Pre-VLA's on two counts: ours gates
\emph{escalation} (handing control to a separate policy) rather than verification
activation within the same policy, and ours adds the explicit budget cap $\Kmax$ that
bounds the escalation tail and keeps ``targeted'' from collapsing into always-strong. We
do not claim the gate is wholly novel.

\paragraph{Trigger count versus duty cycle (what the cap bounds).} Three quantities must
be kept distinct. The budget cap $\Kmax$ bounds the number of \emph{gate fires} (trigger
events) per episode, not the number of strong-policy steps. Because each fire hands control
at the next chunk boundary and holds the stronger policy for at least $k{=}3$ chunks of
length $H{=}10$ before de-escalation is considered, one fire activates the stronger policy
for tens of steps. The realized \emph{duty cycle}, the fraction of executed steps on which
the stronger policy is active, is therefore an emergent consequence of the fire pattern,
the hold, and early task termination; on the confirmatory run it is $0.38$ (step-weighted,
$n{=}700$). The budget-matched controls (C, D; \S\ref{sec:method-escalation}) are matched
to B on this realized strong-step budget and its temporal distribution, not merely on the
trigger count, so a B-over-control win cannot be bought with extra stronger-policy compute.
Finally, the $\leq 30\%$ early window is the probe's \emph{evaluation/precondition} readout
(where we measure AUROC), not a runtime upper cutoff on firing: at deployment the early-harm
guard suppresses switches before $\tmin{=}0.20\,T$, while the conformal trigger may fire
later if the score crosses $\delta_t$, and such late fires are counted as misses for the
early-warning precondition even though the controller still acts on them.

\subsection{Escalation handoff and the four-arm factorial}
\label{sec:method-escalation}

\begin{figure}[htbp]
  \centering
  \includegraphics[width=\linewidth]{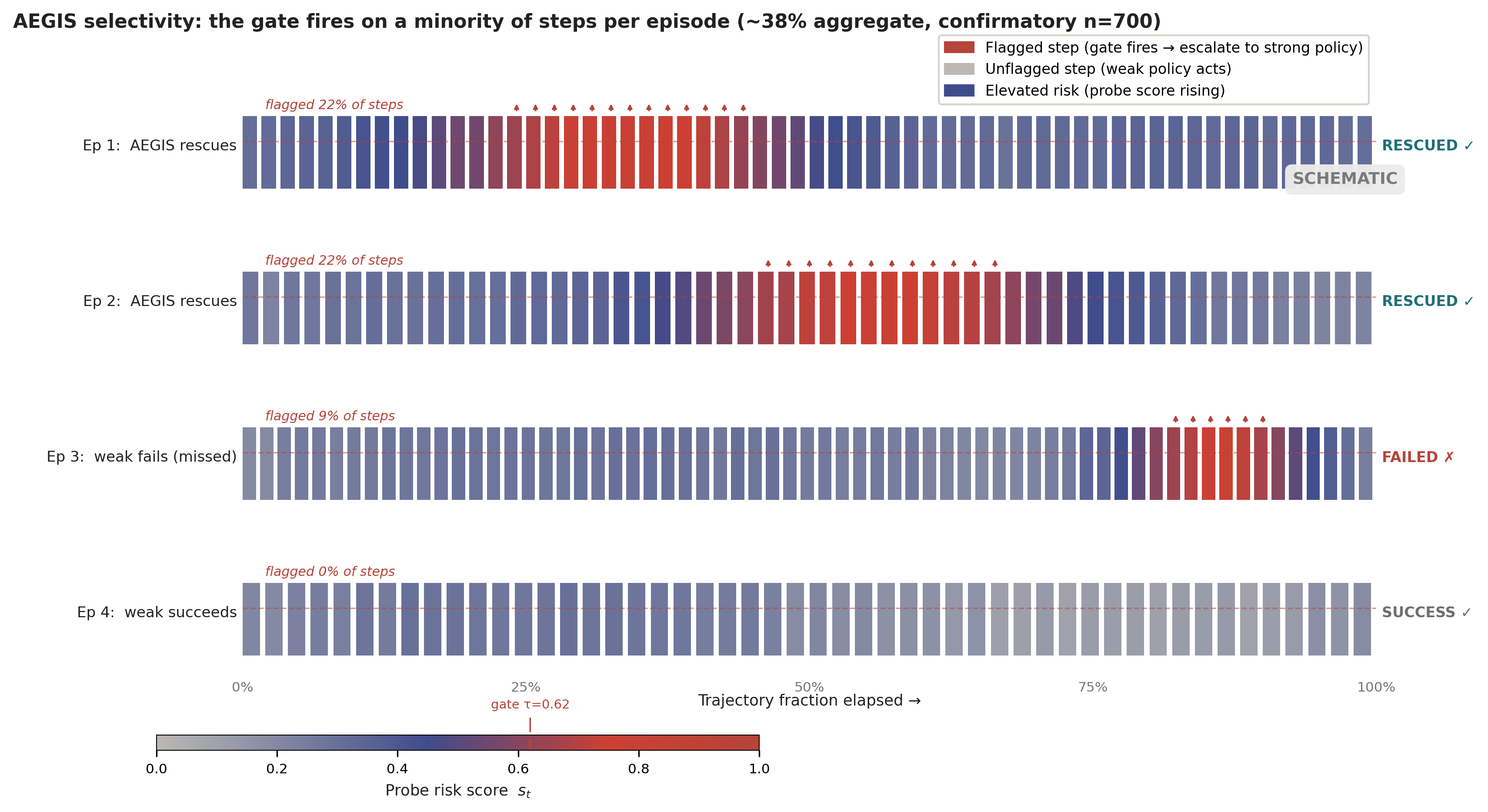}
  \caption{\textbf{Selectivity, per step.} Each row is one episode; bar colour encodes the
  per-step probe risk score $\sscore_t$ (grey low, red high), and red markers show the steps
  where the gate fires and control hands to the stronger policy. AEGIS escalates only a small
  fraction of steps per episode (${\sim}38\%$ in the confirmatory $n{=}700$ run) rather than running the
  stronger policy throughout; the figure also shows a late-detection miss (gate fires only
  past $80\%\,T$) and a clean weak-only success that is never escalated.
  \emph{(Schematic risk profiles; the escalated-step fraction is the measured confirmatory
  $n{=}700$ value.)}}
  \label{fig:trajectory-strip}
\end{figure}

\paragraph{Chunk-boundary handoff, hold, and hysteretic de-escalation.}
Escalation is realized only at action-chunk boundaries. When a step clears the gate and
budget ($\Esc_t{=}1$), control transfers from $\pi_w$ to $\pi_s$ at the next chunk
boundary ($H = 10$ on the executed checkpoints; \S\ref{sec:deviations}), which keeps the switch aligned with the policies' native action
granularity and avoids cutting into a partially-executed chunk. Figure~\ref{fig:trajectory-strip} shows this gating per step across several episodes. Once engaged, $\pi_s$ is
\emph{held} for a minimum of $k = 3$ chunks before any return is considered, and
de-escalation back to $\pi_w$ uses hysteresis: the controller returns to the weak policy
only after the score has stayed below the (lower) de-escalation threshold for the hold
window, so the system does not chatter between policies on a borderline signal. The two
policies do not in general agree on the in-flight action chunk, so a handoff risks a
kinematic discontinuity. Holding the handoff to chunk boundaries (above) is our first
guard. As a future extension we plan to condition the incoming policy's first chunk on the
committed tail of the outgoing chunk in the style of real-time chunking (RTC) inpainting,
so the executed trajectory is continuous in action space. The executed pilot uses the
chunk-boundary handoff without RTC inpainting.

\paragraph{In-process, single-container design (security rationale).}
Both policies are held warm in a \emph{single} process inside one container, so a handoff
is a function call rather than a network round-trip. This is partly a latency choice and
partly a security one: we deliberately never instantiate the framework's networked
\texttt{PolicyServer} path, which is subject to a pickle-deserialization remote-code
execution vulnerability (CVE-2026-25874). Keeping both frozen policies co-resident in one
container removes that attack surface entirely while letting the stronger policy stay warm
(it fits in ${\sim}9.5$\,GB of VRAM, so a single mid-range accelerator suffices). The
weak/strong pair is \texttt{smolvla\_libero} ($450$M) escalating to
\texttt{pi05\_libero\_finetuned} ($4.14$B) \citep{pi052025, pi02024}. We write the per-episode
cost as $\mathrm{Cost}(\text{AEGIS}) \approx C_w + C_{\mathrm{probe}} + \rho\,C_s$ versus
$\mathrm{Cost}(\text{always-strong}) = C_s$, where $C_w, C_s$ are the per-step weak and
strong forward-pass costs, $C_{\mathrm{probe}}$ is the negligible small-MLP probe read, and
$\rho{=}0.38$ is the strong-policy duty cycle. The weak forward pass is retained across the
episode to supply the probe score that governs triggering and hysteretic de-escalation, so
$C_w$ is paid throughout and the stronger policy runs in addition on the $\rho$ fraction of
steps. We report this as a parameter-count-anchored \emph{schematic}
(Fig.~\ref{fig:cost-wedge}), not measured wall-clock, and we therefore speak of a
\emph{matched strong-policy duty} between B and its controls rather than ``the same
compute.'' The design point is that selective escalation is cheaper than running the
stronger policy on every step.

\paragraph{The four experimental arms.}
We evaluate a four-arm factorial, all paired by common random numbers and keyed by
$(\text{task}, \text{seed}, \text{arm})$ via a Philox counter-based generator so that
every arm sees the identical $(\text{task}, \text{seed}, \text{init\_state})$ tuple and
only the intervention policy varies:
\begin{description}[leftmargin=2.2em, itemsep=2pt, topsep=2pt]
  \item[\textbf{A. Weak-only}] the weak policy runs the whole trajectory (baseline floor).
  \item[\textbf{B. Targeted}] signal-gated escalation of flagged steps to the stronger
        policy (\emph{the method}).
  \item[\textbf{C. Budget-matched-blind}] the same escalation \emph{count} and the same
        step-index-fraction CDF as B (conditioned on episode-length bucket), but with the
        escalated steps chosen \emph{blind} of any signal, which isolates selectivity from
        raw compute.
  \item[\textbf{D. Random-trigger placebo}] the same per-step fire rate as B, with steps
        chosen uniformly at random, which isolates the information in the signal.
\end{description}
We additionally report an \textbf{always-strong} ceiling (the stronger policy on the whole
trajectory; an upper-bound reference, not a primary contrast). As required comparison
baselines on the same rollouts we re-implement a HELM-style rollback-to-checkpoint plus
goal-conditioned replan-and-recover \emph{within the weak policy} ($\Rmax = 3$),
budget-matched to B's extra compute \citep{helm2026}, and a SAFE-style detect-and-halt
with no recovery as the detection-only reference \citep{safe2025}. We follow the reporting
conventions of the HELM and SAFE evaluations for comparability on LIBERO
\citep{helm2026, safe2025}.

\paragraph{A second strong policy, for generalization.}
To show that the effect is a property of escalating to a stronger
\emph{separate} policy and not of one lucky weak/strong pair, the main factorial repeats
the targeted arm B with the escalation \emph{target} swapped from $\pi_{0.5}$ to NVIDIA's
GR00T~N1.x (the official LIBERO checkpoint, run LeRobot-native), holding the deployed weak
policy, signal, and threshold fixed \citep{groot2503.14734}; GR00T is an escalation target
only, never a signal source, and this generalization arm is a robustness result, not
one of the make-or-break primary contrasts, which stay on the $\pi_{0.5}$ pair.

\subsection{Causal identification: controller vs. difficulty proxy}
\label{sec:method-causal}

The purpose of this paper is a \emph{causal} claim, that escalating where the signal
fires recovers tasks \emph{because} of where it escalates, so we state the estimand,
the identifying contrasts, and the condition under which the claim is falsified, all
before any data exists.

\paragraph{Estimand.}
The target estimand is the recovered-task-rate, conditional on the subset of trajectories
the weak policy alone fails:
\[
  \RTR_B \;=\; \Pr\!\big[\,\text{task succeeds under arm } B \;\big|\;
             \text{weak-only arm } A \text{ fails on that }(\text{task},\text{seed})\,\big].
\]
The conditioning event is essential: a method that escalates can only \emph{recover} a
trajectory that would otherwise have failed, so the relevant population is the $A$-failing
subset, not all episodes. Because every arm is run under common random numbers on the same
$(\text{task},\text{seed},\text{init\_state})$ tuple, ``$A$ fails'' is observed on the
\emph{same} trajectory whose arm-$B$ outcome we score, and the contrast of interest is the
within-pair difference in success on those discordant trajectories.

\paragraph{Why $B > A$ alone does not identify the effect.}
Arm $B$ spends strictly more compute than arm $A$ (it runs the stronger policy on some
steps), and the steps it escalates are, by construction, the steps the signal finds
alarming, which are also, on average, the harder steps. A raw improvement of $B$ over
$A$ is therefore consistent with at least two non-causal explanations: (i) a
\emph{raw-compute} artifact, in which any extra application of the stronger policy would
help regardless of where it is placed; and (ii) a difficulty-proxy artifact,
in which the signal merely indexes which trajectories are hard and the apparent gain is a
re-description of difficulty rather than an effect of selective intervention. The
intervention-paradox result makes the danger concrete: prediction accuracy alone licenses
no claim about a controller \citep{interventionparadox2026}. We therefore design two
controls that hold the confounds fixed and vary only the thing we claim is causal:
\emph{where} the compute is spent.

\paragraph{The two identifying controls.}
Arm $C$ (budget-matched-blind) escalates the \emph{same number} of steps as $B$, drawn so
that its step-index-fraction CDF matches $B$'s within each episode-length bucket, but
chooses \emph{which} steps blind of the signal. $C$ thus equalizes total stronger-policy
compute and its temporal distribution; the only thing it removes is the signal's
selectivity. A win of $B$ over $C$ cannot be explained by raw compute, because compute is
held equal. It can only be explained by escalating the \emph{right} steps. Arm $D$
(random-trigger placebo) fires at the \emph{same per-step rate} as $B$ but on uniformly
random steps; it removes the information in the signal while preserving its intensity. A
win of $B$ over $D$ shows the trigger carries real predictive information rather than
firing as a constant-rate process. Holding both contrasts \emph{within difficulty strata}
(\S\ref{sec:design}) closes the difficulty-proxy loophole: if $B$ beats $C$ and $D$ even inside
the medium-difficulty stratum where every arm faces comparably hard trajectories, the gain
cannot be a re-description of cross-trajectory difficulty. Probing studies that establish
a separating signal typically rely on label-shuffle and temporal controls rather than a
budget-matched and a rate-matched intervention control, which is exactly why such evidence
alone cannot make the causal selectivity claim we make here.

\paragraph{Falsification (the kill / null condition).}
The claim is falsified, and the paper is reported honestly as a characterization rather
than a controller result, if either of the following holds. (K1) The early-window probe
AUROC sits at chance, so there is no anticipatory signal to act on. (K2) The $\RTR$ gain
of $B$ does \emph{not} survive difficulty stratification \emph{and} does not beat $C$
\emph{and} does not beat $D$, in which case the signal is a difficulty proxy rather than
a controller, and we say so. Two further pre-committed conditions bound scope: (K3) if the
escalation fraction exceeds $50\%$, ``targeted'' has collapsed into always-strong and the
cost argument dies; and (K4) if prior art appears that also escalates flagged steps to a
stronger \emph{separate} policy with a recovery metric, the novelty axis is occupied and
we reassess. These conditions are named here, before data, so that the null that would
sink the claim is not negotiable after the fact.

\FloatBarrier
\section{Experimental Design (Pre-Registered)}
\label{sec:design}

\emph{This analysis plan was frozen before any experimental result existed.} Deviations
discovered after seeing data are reported in the dedicated ``Deviations from
Pre-Registration'' subsection (\S\ref{sec:deviations}) below.

\paragraph{Primary endpoint and conditional estimand.}
The headline is the recovered-task-rate,
$\RTR_B = \Pr[\,\text{task succeeds under B} \mid \text{weak-only arm A fails on that }
(\text{task},\text{seed})\,]$. This is conditional on the A-failing subset, whose size
is the number of A-failures rather than the total episode count; we power on this
conditional subset, not on aggregate episode count. The unit of paired evidence is a
\emph{discordant pair}: a $(\text{task},\text{seed})$ tuple where two arms differ in
success. Common random numbers make A/B/C/D paired at the trajectory level: every arm
sees the same $(\text{task},\text{seed},\text{init\_state})$ tuple, and only the
intervention policy varies, which is what licenses the paired tests below.

\paragraph{Primary contrasts and multiplicity.}
The primary family of exactly three contrasts is tested under Holm--Bonferroni \citep{holm1979} at
family-wise $\alpha = 0.05$: (1) $B > A$ (recovery exists at all), (2) $B > C$ (beats
budget-matched-blind spend, isolating selectivity from raw compute), and (3) $B > D$
(beats the random-trigger placebo, isolating the information in the signal). All three are
\emph{one-sided} (each hypothesis is directional, $B$ above the named control). Contrasts
(2) and (3) use an exact paired McNemar test \citep{mcnemar1947} on the discordant pairs of
the conditional ($A$-failing) subset. Contrast (1) is a special case: on the $A$-failing
pool $A$ succeeds on zero trajectories by construction, so the $B{>}A$ comparison has no
$A$-win discordant cell and the McNemar test reduces to an exact one-sided binomial (sign)
test on the count of $A$-failing trajectories that $B$ recovers; we report it as such.
Each contrast is reported alongside a paired bootstrap \citep{efron1993bootstrap} of
$B = 10{,}000$ resamples that resamples \emph{whole trajectories}, never individual steps,
to give a $95\%$ confidence interval on the $\RTR$ difference. The method ``wins'' a
contrast if and only if the Holm-adjusted $p < 0.05$ \emph{and} the $95\%$
whole-trajectory bootstrap CI excludes $0$ in the predicted direction; both conditions are
required. The secondary contrast
$B > \text{HELM-baseline}$ \citep{helm2026} is a separate family controlled at
BH-FDR $= 0.10$ and is reported as secondary. Marginal success rates use Wilson intervals
and small cells use Clopper--Pearson; per-task results are shown as forest plots with a
pooled diamond.

\paragraph{Within-stratum requirement.}
Difficulty strata are terciles of weak-only base success per task, frozen from a
$50$-rollout pilot: easy (top tercile), medium (middle, weak base success $20$--$70\%$,
maximum recovery headroom), and hard (bottom tercile). For the headline claim, the RTR
gain of B over each control must be $> 0$ \emph{within every} stratum that clears the
discordant-pair floor, with special attention to the medium stratum. A win that exists
only in the pooled estimate but vanishes or reverses inside a stratum is reported as
\emph{not} supporting the causal claim.

\paragraph{Power and hard floors.}
The sample size derives from one effect and one power target. The pre-stated
method-relevant effect is a $10$ percentage-point gap in conditional $\RTR$ between $B$ and
the relevant control. The power target is $80\%$ for the $B>C$ gap, the hardest of the
three primary contrasts ($B>A$ is easier and $B>D$ easier still). At $80\%$ McNemar power
for a $10$pp shift in discordant proportions, this requires on the order of $250$--$400$
discordant/failing pairs, so we pre-commit to a hard $250$ floor and a target of $400$.
Weak chained-LIBERO base success is estimated near $0.5$, so roughly half of episodes
enter the $A$-failing subset. Reaching that many failing pairs then implies on the order of
$700$--$800$ episodes per arm, and we pre-commit to $\geq 700$ episodes per headline arm for
cells $A/B/C/D$. A binding hard floor of $\geq 20$ discordant pairs \emph{per primary
contrast per difficulty stratum} also applies. Any stratum below that floor is reported with
exact-binomial / mid-$P$ only and is never cited as within-stratum confirmation. The
$50$-rollout weak-only pilot fixes the difficulty terciles, the base-rate that converts the
discordant-pair target into the final episode count, and the conformal thresholds; the
pilot's numbers feed these counts but never \emph{relax} any floor. Episode counts are
fixed in advance and there is no optional stopping on primary $p$-values.

\paragraph{Kill criteria.}
We downgrade to a characterization paper if any of: (K1) early-window probe AUROC sits
at chance; (K2) the RTR gain does not survive stratification \emph{and} does not beat C
\emph{and} does not beat D, in which case it is a difficulty thermometer, not a controller,
and we publish that honestly; (K3) the chained escalation fraction exceeds $50\%$, so
``targeted'' collapses into always-strong and the budget claim dies; or (K4) a prior-art
paper appears that also escalates flagged steps to a stronger separate policy with a
recovery metric.

\subsection{Deviations from Pre-Registration}
\label{sec:deviations}
We log every departure of the executed pilot from the frozen plan, with the reason and
the scope of its effect.
\begingroup\sloppy
\begin{description}[leftmargin=1.8em, itemsep=3pt, topsep=3pt]
  \item[\textbf{Action-chunk horizon $H$: $50 \to 10$.}] The pre-registration described
        $H = 50$; the executed checkpoints emit and integrate action chunks of
        \texttt{n\_action\_steps} $= 10$, so we set $H = 10$ to match the policies' native
        granularity. Effect: finer escalation granularity; no change to the estimand or to
        any contrast definition.
  \item[\textbf{Suite: LIBERO-Spatial.}] The main factorial runs on
        \emph{LIBERO-Spatial} (10 tasks), the regime selected on a four-suite pilot sweep as
        the in-band difficulty band where the weak policy fails often enough to leave
        recovery headroom (weak base success $\approx 0.33$, strong ceiling $\approx 0.95$,
        recovery headroom $\approx 1.0$). The pre-registration named LIBERO-Long and
        chained-LIBERO as candidate long-horizon suites; the pilot sweep showed those bands
        either out of the $20$--$70\%$ weak-failure window or with lower headroom, so we
        committed to LIBERO-Spatial for the confirmatory factorial. Extending the same
        protocol to chained, long-horizon suites is the natural next step (future work).
  \item[\textbf{Probe source: vision-encoder hook (bug) $\to$ live action-expert
        layer~15 \texttt{o\_proj} ($d{=}720$).}] An initial implementation hooked a
        vision-encoder layer and captured a feature that did not vary with the rollout,
        yielding chance-level prediction (AUROC $0.50$). We diagnosed this as a
        hook-placement bug (the captured tensor was a frozen cached feature, not a live
        activation), and moved the probe to the action expert's self-attention output
        projection at layer~15, read live during rollout. All reported AUROC numbers use
        the corrected live probe. This is a bug fix, not a post-hoc head search: the
        layer was fixed before scoring and the std-over-steps $> 0.05$ check guards
        against re-introducing a frozen feature.
  \item[\textbf{Two-stage evidence: gate pilot then confirmatory factorial.}]
        The pre-registration pre-commits to $\geq 700$ episodes per arm and a
        discordant-pair floor for the within-stratum analysis. The go/no-go gate (Phase-D)
        was powered at $56$ common-random-number keys per arm ($41$ in the A-failing
        conditional subset); its confidence intervals are wide \emph{by design}. The
        confirmatory factorial reported in the main results scales the same protocol on
        LIBERO-Spatial to $\geq 700$ episodes per arm across the full
        $10\times 70$ task-by-seed grid and the full arm set (A weak-only,
        B targeted cascade, C budget-matched-blind, D random-trigger, HELM same-policy
        rollback, always-strong ceiling, and the GR00T~N1.7 cross-family generalization
        arm), supplying the within-stratum confirmation at the pre-registered floor.
\end{description}
\endgroup
The abstract and method report the LIBERO-Spatial confirmatory factorial as the headline
evidence, with the Phase-D gate retained as the pre-committed go/no-go decision that
justified scaling to the full run.

\FloatBarrier
\section{Results}
\label{sec:results}

\begin{figure}[htbp]
  \centering
  \includegraphics[width=\linewidth]{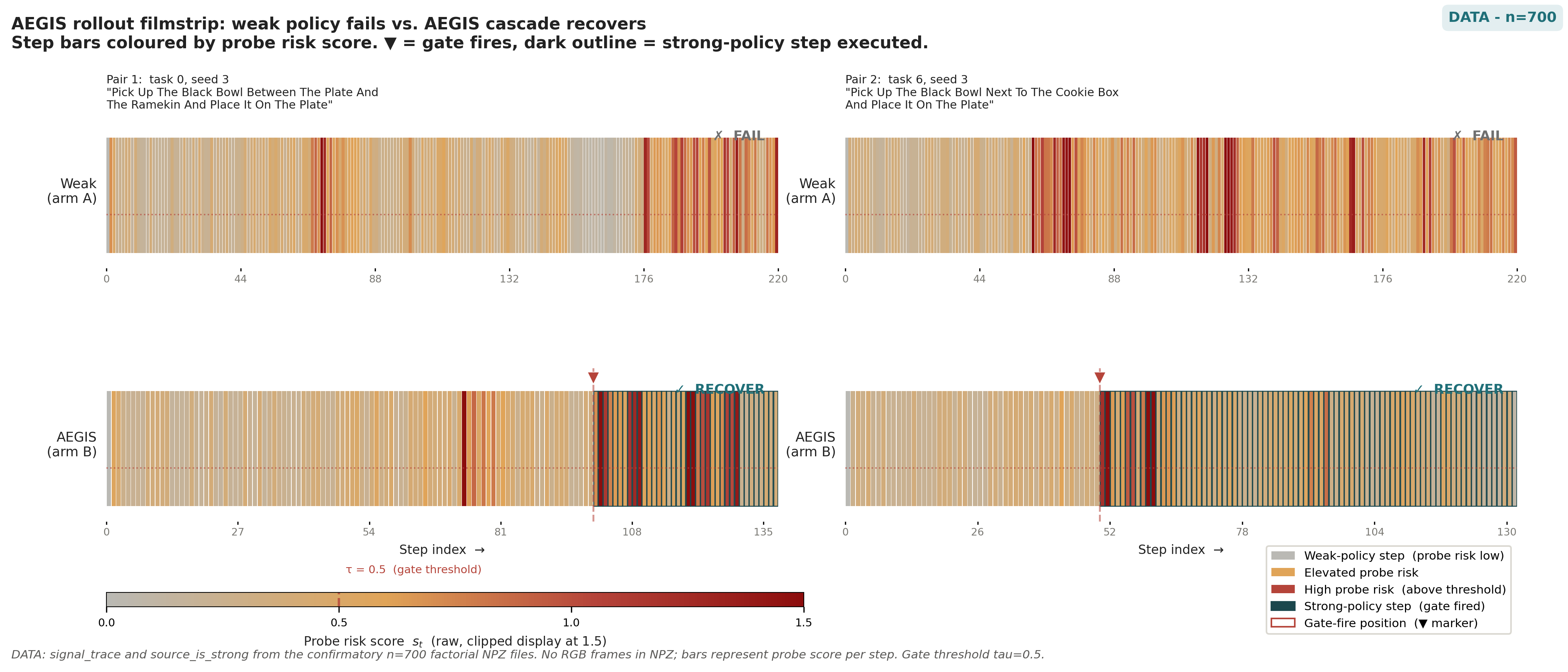}
  \caption{\textbf{What the controller does, on two real episodes.} Each panel is one
  common-random-number key (same task, same seed, same initial state) run under two arms.
  \emph{Top rows (Weak, arm A):} the deployed weak policy alone; per-step probe risk
  ($\sscore_t$) climbs and the trajectory ends in failure. \emph{Bottom rows (AEGIS, arm B):}
  on the identical key, the gate fires ($\blacktriangledown$) in the early window and control
  hands to the stronger policy (dark, outlined steps), recovering the task. Bars encode the
  per-step probe score from the logged \texttt{signal\_trace}; strong-policy steps come from
  the logged \texttt{source\_is\_strong} mask, and gate-fire positions from the recorded
  hand-off steps. \emph{(Real confirmatory $n{=}700$ rollout data; the NPZ logs carry per-step scores and
  policy-source masks but not RGB frames, so steps are shown as a risk-coloured strip rather
  than rendered images.)}}
  \label{fig:filmstrip}
\end{figure}

The headline result is the confirmatory factorial on LIBERO-Spatial
(Table~\ref{tab:confirm-rtr}): the full $10\times 70$ task$\times$seed grid at
$\geq 700$ common-random-number episodes per arm, with the within-stratum
confirmation at the pre-registered discordant-pair floor. A small Phase-D go/no-go
pilot ($n=56$ keys per arm) established directionality and triggered this
pre-registered run; we do not use it for the headline claim, and its plots sit in
Appendix~\ref{app:pilot}. Beyond the four primary arms (A, B, C, D), the factorial
collects the HELM same-policy rollback arm (re-running the \emph{same} weak policy on
flagged steps), the always-strong recovery ceiling, and the GR00T~N1.7 arm that
escalates to a different policy family to test cross-family generalization. The run
reported here is complete: all $700$ cells carry the five core arms
($A,B,C,D$, always-strong) under a single-host pairing constraint, with HELM observed
on $692$ cells and GR00T on all $700$. Figure~\ref{fig:filmstrip} first shows, on two
real common-random-number keys, what the controller actually does: the weak policy
alone fails while AEGIS fires the gate early and recovers the same task.

\begin{figure}[htbp]
  \centering
  \includegraphics[width=0.92\textwidth]{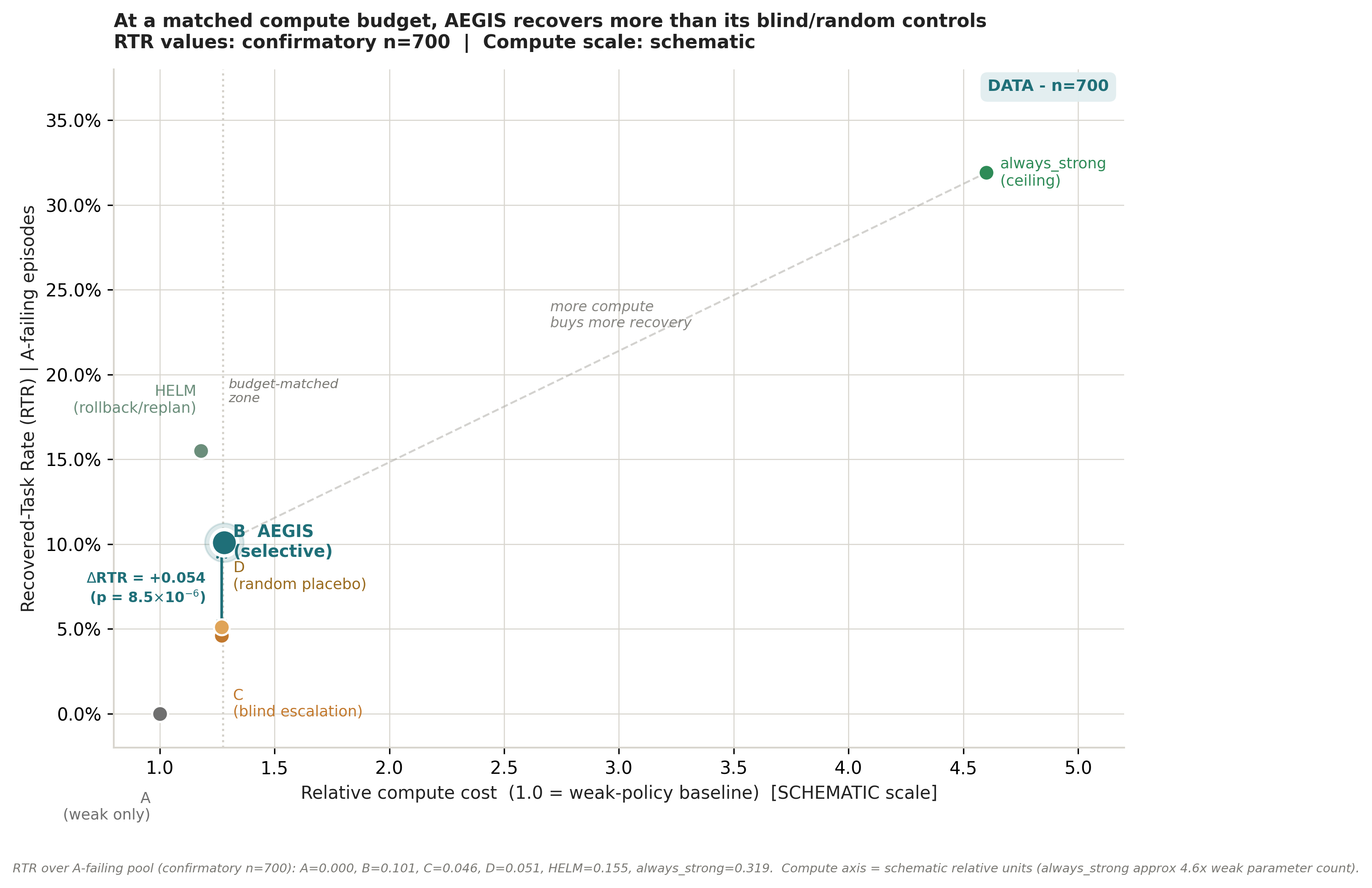}
  \caption{\textbf{Selectivity, not spend, is the lever.} Each arm is placed by relative
  compute cost (horizontal; weak-policy baseline $=1.0$) against recovered-task rate
  (vertical, confirmatory $n{=}700$ data). The budget-matched controls C and D sit in the same cost
  bracket as AEGIS (B) but far below it, while always-strong reaches near-ceiling recovery
  only at $\approx 4.6\times$ the compute. At the shared, near-baseline compute budget,
  AEGIS recovers the most of any arm in that bracket; spending more compute (HELM, GR00T,
  always-strong) buys more recovery, so the claim is selectivity at a fixed budget, not
  raw dominance over the higher-compute arms. \emph{(RTR values are measured confirmatory
  $n{=}700$ data; the compute axis uses schematic relative units anchored to parameter
  counts.)}}
  \label{fig:quadrant}
\end{figure}

\begin{table}[htbp]
  \centering
  \caption{\textbf{Confirmatory factorial} on LIBERO-Spatial: recovered-task-rate (RTR) and
  marginal success by arm, with paired contrasts ($\geq 700$ episodes per arm; full
  $10\times 70$ task$\times$seed common-random-number grid; horizon $H=10$). RTR is
  conditional on the weak-only~(A) failing subset. $p$-values are Holm-adjusted over the
  three primary contrasts (B${>}$A, B${>}$C, B${>}$D) using one-sided exact paired tests on
  the conditional ($A$-failing) pool; the always-strong arm bounds the recovery ceiling.
  The discordant column gives (B-win\,:\,arm-win) pairs entering each exact test: for $B{>}A$
  the arm-win cell is zero by construction (\S\ref{sec:design}), so that test is a one-sided
  binomial on $B$'s recoveries; $B{>}C$ and $B{>}D$ are McNemar on their discordant pairs.
  The two secondary reference arms (HELM, GR00T) report the unadjusted exact $p$ on the same
  pool; they are reference foils, not primary contrasts, and B is expected to sit below them.}
  \label{tab:confirm-rtr}
  \resizebox{\textwidth}{!}{%
  \begin{tabular}{l c c c c c}
    \toprule
    Arm & Marginal success & Conditional RTR & Contrast (B $-$ arm) & Discordant (B:arm) & Exact $p$ \\
    \midrule
    A. Weak-only             & $0.077$ & $0.000$ & $+0.101$ & $65\!:\!0$  & $5.2\times10^{-11}$ \\
    B. Targeted (method)     & $0.156$ & $0.101$ & reference & --         & -- \\
    C. Budget-matched-blind  & $0.096$ & $0.046$ & $+0.054$ & $65\!:\!23$ & $8.5\times10^{-6}$ \\
    D. Random-trigger        & $0.110$ & $0.051$ & $+0.050$ & $49\!:\!17$ & $1.0\times10^{-4}$ \\
    HELM (same-policy roll.) & $0.205$ & $0.155$ & $-0.055$ & $63\!:\!96$ & $6.9\times10^{-3}$ \\
    \midrule
    GR00T~N1.7 (cross-family)& $0.191$ & $0.155$ & $-0.054$ & --          & $4.0\times10^{-3}$ \\
    Always-strong (ceiling)  & $0.360$ & $0.319$ & $-0.218$ & --          & -- \\
    \bottomrule
  \end{tabular}%
  }
\end{table}

\begin{table}[htbp]
  \centering
  \small
  \caption{\textbf{Recovery versus disruption} on the confirmatory grid ($n{=}700$;
  $54$ trajectories the weak policy alone succeeds on, $646$ it fails). Because escalation
  can also derail a trajectory that would have succeeded \citep{interventionparadox2026},
  we account for both effects of each arm against the weak-only outcome. Selectivity shows
  up as the recover-to-disrupt ratio: AEGIS recovers the most failures while disrupting the
  fewest successes; blind escalation disrupts the most for the least recovery. Counts derive
  from the same paired discordant cells as Table~\ref{tab:confirm-rtr} (B's disruption count
  equals the $B{>}A$ arm-win cell).}
  \label{tab:harm}
  \begin{tabular}{l c c c c c}
    \toprule
    Arm & A-failures recovered & A-successes preserved & A-successes disrupted & recover\,:\,disrupt & net success \\
    \midrule
    A. Weak-only            & $0/646$  & $54/54$ & $0/54$  & --     & $54$ \;\,($7.7\%$) \\
    B. Targeted (AEGIS)     & $65/646$ & $44/54$ & $10/54$ & $6.5$  & $\mathbf{109}$ ($\mathbf{15.6\%}$) \\
    C. Budget-matched-blind & $30/646$ & $37/54$ & $17/54$ & $1.8$  & $67$ \;\,($9.6\%$) \\
    D. Random-trigger       & $33/646$ & $44/54$ & $10/54$ & $3.3$  & $77$ \;\,($11.0\%$) \\
    \bottomrule
  \end{tabular}
\end{table}

\begin{figure}[htbp]
  \centering
  \includegraphics[width=0.8\textwidth]{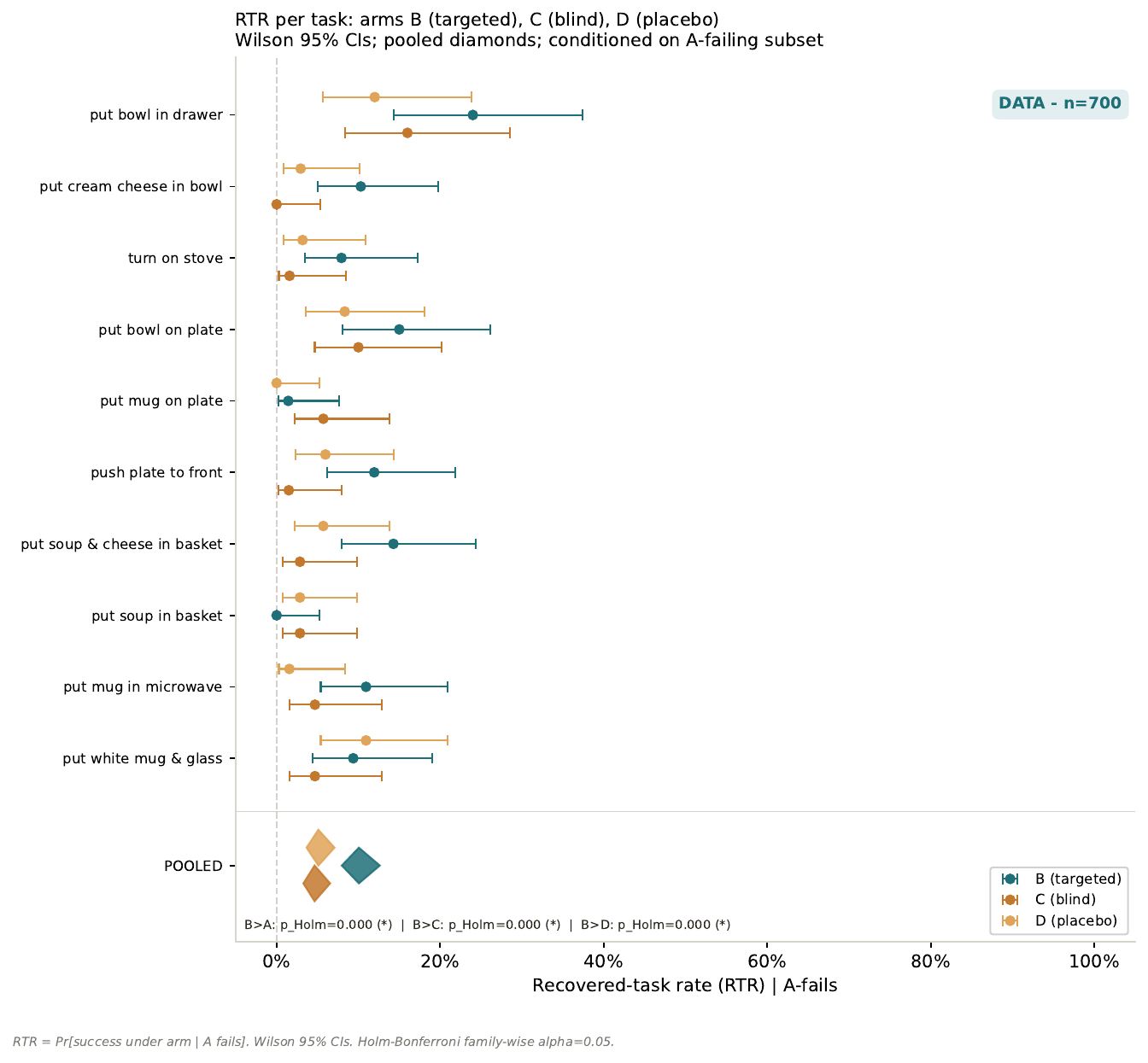}
  \caption{\textbf{Per-task recovered-task rate, conditioned on the A-failing subset.}
  Confirmatory factorial ($n{=}700$ CRN cells): for each of the $10$ LIBERO-Spatial tasks,
  conditional RTR for arms B (targeted), C (budget-matched-blind), and D (random-trigger),
  with Wilson $95\%$ intervals. The bottom POOLED row shows the across-task pooled diamonds
  (targeted~B near $0.10$, the budget-matched controls~C and~D near $0.05$). All three
  primary contrasts ($B{>}A$, $B{>}C$, $B{>}D$) clear Holm-adjusted significance on the
  pooled estimate. The per-stratum (easy/medium/hard) version of these contrasts appears in
  Fig.~\ref{fig:hazard} at the pre-registered $\geq 20$-discordant-pair floor.}
  \label{fig:forest}
\end{figure}

\begin{figure}[htbp]
  \centering
  \includegraphics[width=0.95\textwidth]{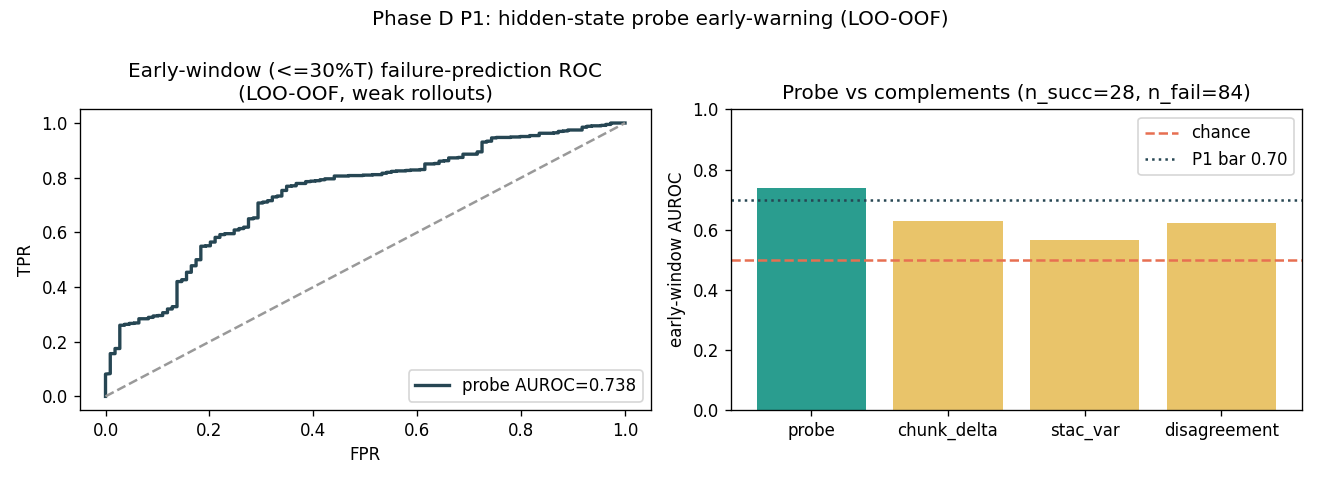}
  \caption{Early-window ($t\leq 0.30\,T$) failure-prediction performance of the
  hidden-state probe on the Phase-D pilot (leave-one-out out-of-fold; $28$ success /
  $84$ failure episodes). \emph{Left:} ROC of the live action-expert probe, early-window
  AUROC $=0.738$ (the confirmatory run re-estimates this at $0.764$ over $n{=}2{,}792$
  episodes, clearing the $\geq 0.75$ main-run precondition; Fig.~\ref{fig:wedge}).
  \emph{Right:} the probe versus three
  surprise/disagreement complements on the same rollouts; the probe is the only signal
  clearing the $0.70$ pilot bar.}
  \label{fig:auroc}
\end{figure}

\begin{figure}[htbp]
  \centering
  \includegraphics[width=0.95\textwidth]{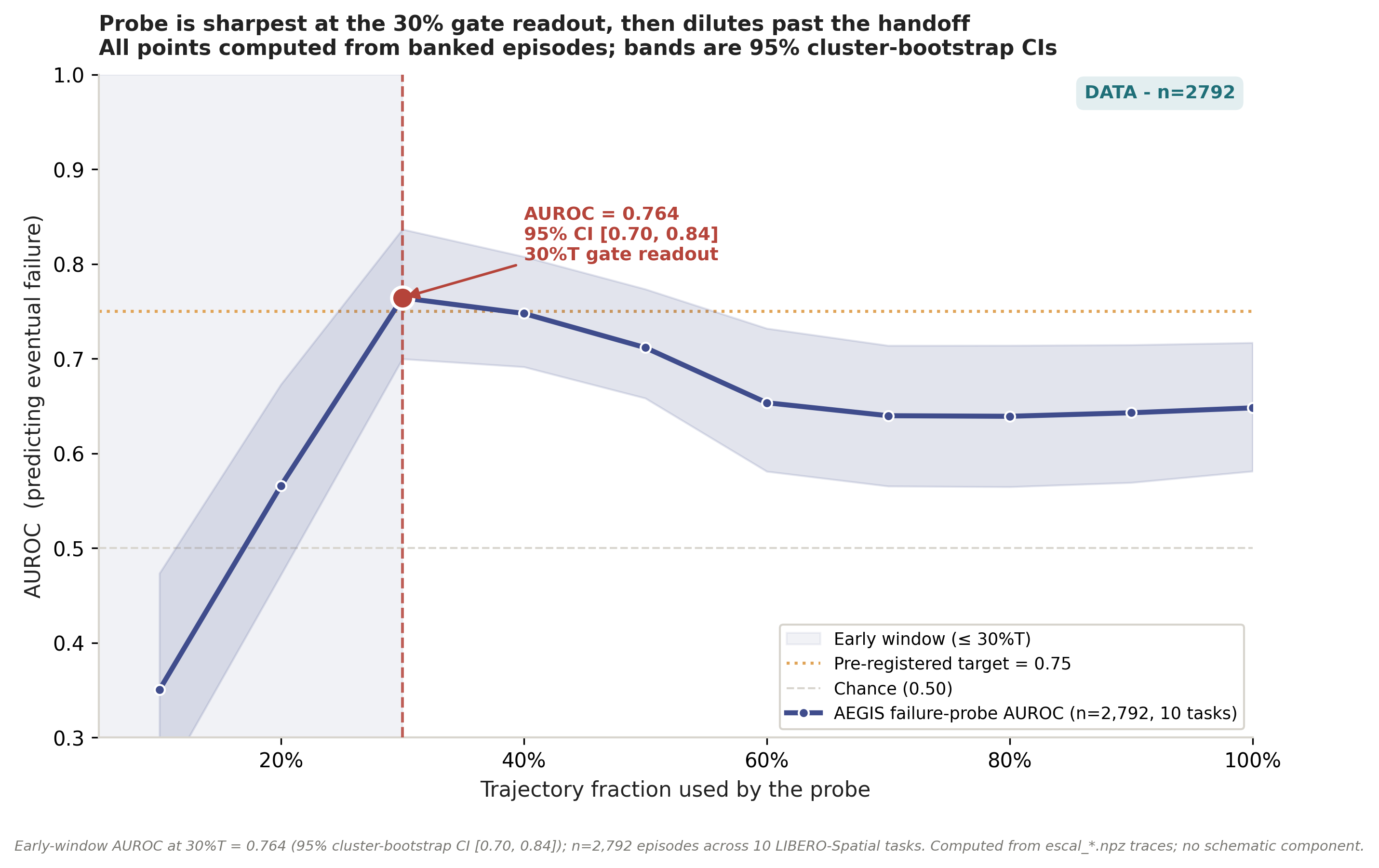}
  \caption{\textbf{The probe is sharpest at the gate readout.} Failure-prediction AUROC of
  the hidden-state probe as a function of the trajectory fraction the probe is allowed to
  read, computed on $n{=}2{,}792$ episodes across $10$ tasks (every point and band is data,
  not a schematic). The headline early-window number is read from the \emph{weak-policy
  path before any handoff}, so the label (eventual failure under the weak policy) and the
  signal are not intervention-contaminated. Discrimination peaks at the $30\%$ gate-readout
  mark (highlighted point: early-window AUROC $=0.764$, $95\%$ cluster-bootstrap CI
  $[0.70,0.84]$, clearing the pre-registered $0.75$ precondition). The decline to the right
  is diagnostic only: it shows what happens when the read window is extended past the point
  where AEGIS would have handed off, mixing in post-switch steps; we do not use those
  extended windows for the precondition. That the signal is most discriminative exactly in
  the window the gate reads, rather than monotonically improving with more steps, is the
  early-warning signature the controller exploits. Bands are $95\%$ cluster-bootstrap
  confidence intervals resampling tasks.}
  \label{fig:wedge}
\end{figure}

\begin{figure}[htbp]
  \centering
  \includegraphics[width=0.78\textwidth]{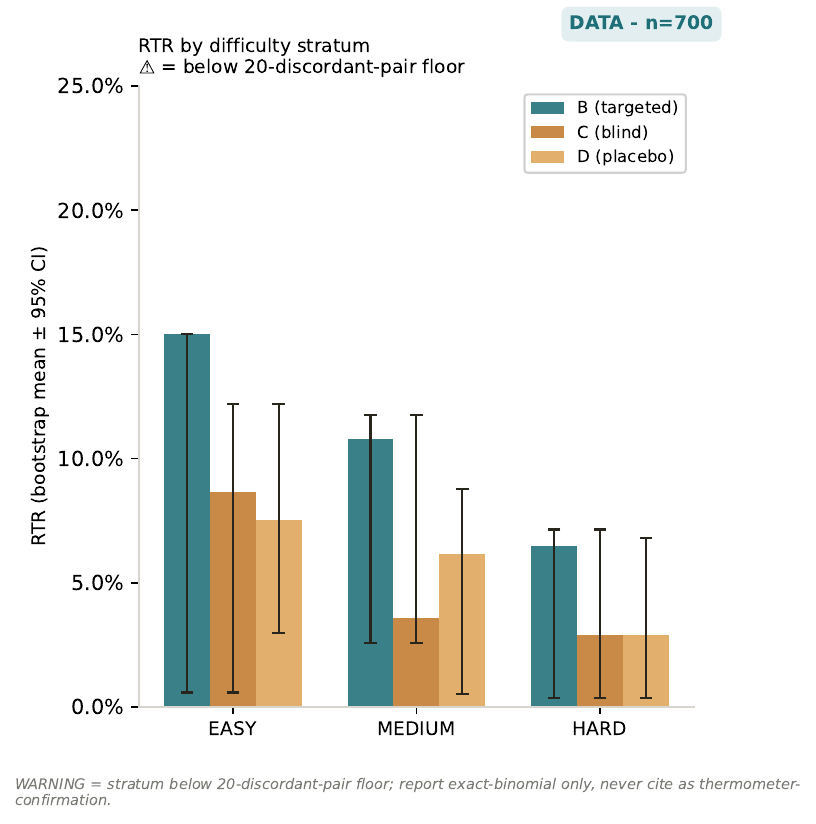}
  \caption{Conditional recovered-task rate (RTR) by difficulty stratum on the
  confirmatory factorial run ($n{=}700$ CRN cells; $n_{A\text{-fail}}{=}646$).
  Point estimates favor targeted escalation~(B) over budget-matched-blind~(C) and
  random-trigger~(D) in all three terciles; the advantage is largest on EASY tasks (where a
  single well-timed escalation suffices) and compresses on HARD tasks, where the $B{-}C$
  interval touches zero (even the strong policy has thin recoverable margin there). Bars are
  bootstrap means with $95\%$ percentile CIs ($n_{\text{boot}}{=}10{,}000$).}
  \label{fig:hazard}
\end{figure}

\paragraph{Confirmatory result ($n{=}700$).}
The full $10\times70$ factorial completed at $n{=}700$ single-host common-random-number
cells ($n_{A\text{-fail}}{=}646$); Table~\ref{tab:confirm-rtr} reports it. All three
primary contrasts clear the pre-registered bar under Holm correction. Arm~B lifts the
conditional recovered-task rate from $0.000$ (weak-only) to $0.101$, beating
budget-matched-blind~(C, RTR $0.046$; $B{-}C=+0.054$, exact-McNemar $p=8.5{\times}10^{-6}$)
and random-trigger~(D, RTR $0.051$; $B{-}D=+0.050$, $p=1.0{\times}10^{-4}$). Every
paired-bootstrap interval excludes zero ($n_{\text{boot}}{=}10{,}000$). At matched cost
the controls sit far below the targeted arm in the recovery-versus-compute plane
(Fig.~\ref{fig:quadrant}). The selectivity signature survives stratification
(Fig.~\ref{fig:hazard}): $B>D$ holds in all three difficulty terciles and $B>C$ holds in
the EASY and MEDIUM bands, with the HARD-band $B{-}C$ interval just touching zero, the
expected attenuation where even the strong policy retains little recoverable margin. The
advantage also holds task by task (Fig.~\ref{fig:forest}). The stronger policy is active on
only $38\%$ of steps (its duty cycle, step-weighted, $n{=}700$ grid) yet B roughly doubles
the recovery of blind or random escalation at matched strong-policy duty. The selectivity
is also visible in the harm accounting (Table~\ref{tab:harm}): B recovers $65$ of $646$
weak-policy failures while disrupting only $10$ of $54$ weak-policy successes, a recover-to-disrupt
ratio of $6.5$, against $1.8$ for blind escalation and $3.3$ for the random trigger. As designed, the cascade does not
beat the full reference policies on raw recovery: HELM (RTR $0.155$) and the cross-family
GR00T~N1.7 arm (RTR $0.155$) both run a strong policy from the first flagged step and sit
above~B, while always-strong bounds the ceiling at $0.319$ on the same pool. That is the
intended trade: at a fraction of the duty cycle, targeted escalation extracts twice the
recovery of the matched-budget controls (Fig.~\ref{fig:cost-wedge}), without claiming to
out-recover an always-on strong policy.

\paragraph{The pilot, demoted.}
The Phase-D pilot was a pre-committed go/no-go check, not a headline. It established the
direction and rough magnitude of the effect on $n=56$ keys per arm and authorized the
confirmatory run; all three primary contrasts cleared their pilot bars there as well. We
report it for completeness in Appendix~\ref{app:pilot} and base no claim on it. Its larger
conditional rates reflect the small, easier pilot pool, not a stronger effect than the
confirmatory $10.1\%$.

\subsection{Robustness to simulator non-determinism}
\label{sec:robustness}
\begin{figure}[htbp]
  \centering
  \includegraphics[width=0.92\textwidth]{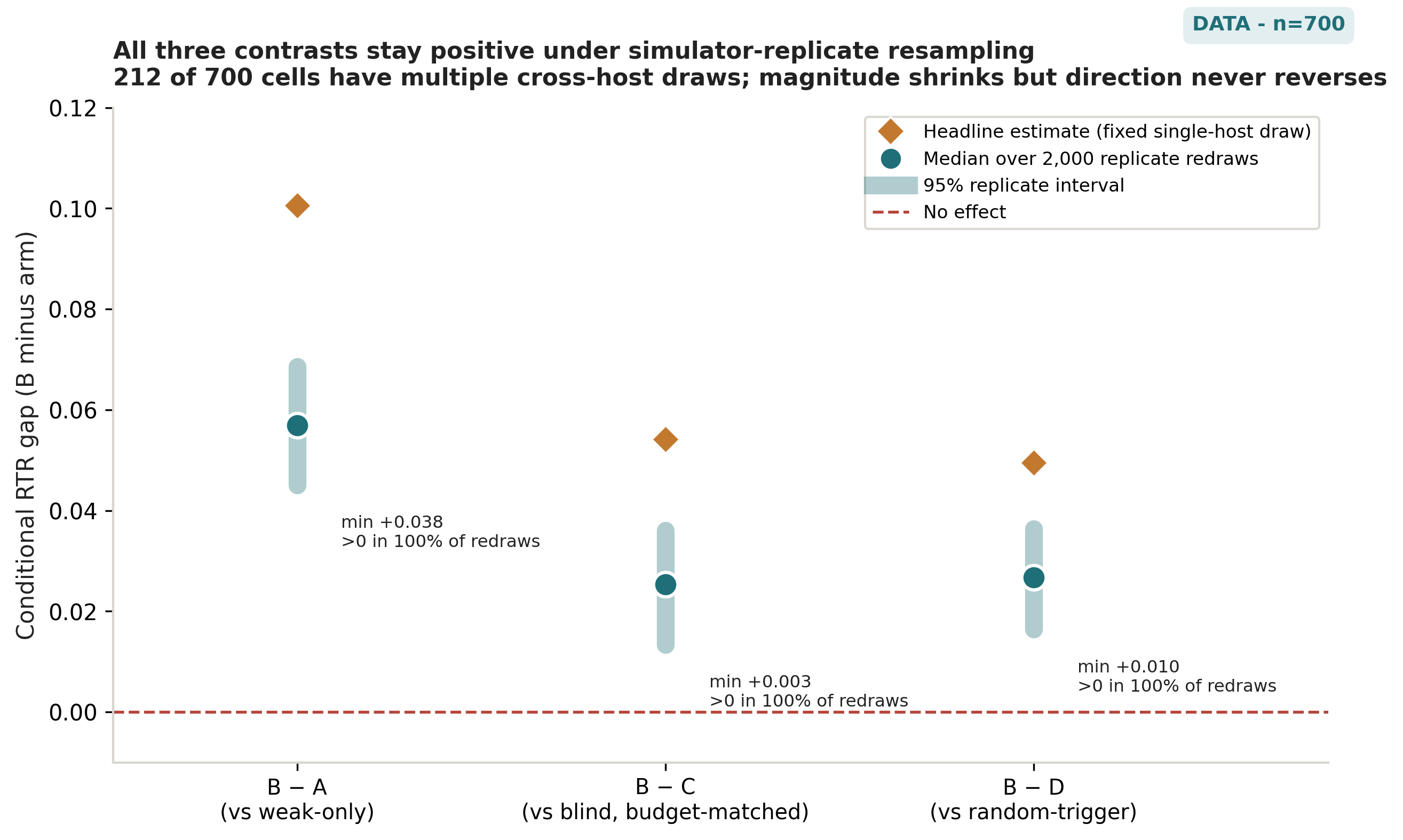}
  \caption{\textbf{Replicate-resampling robustness of the primary contrasts.} For each
  (task,~seed) cell rolled out on more than one host ($212$ of $700$), the simulator is
  not bit-identical, so the success bit can differ across draws. Resampling which
  available single-host-complete draw defines each cell ($2{,}000$ iterations) yields the
  plotted distribution of the three primary RTR gaps. Diamonds are the single-host headline
  estimates; circles and bars are the replicate median and $95\%$ interval. Every
  distribution stays strictly above zero (direction never reverses); the magnitude shrinks
  under adversarial redraws because flipping a borderline cell's weak-arm outcome removes it
  from the A-failing pool (\S\ref{sec:robustness}).}
  \label{fig:robustness}
\end{figure}

The headline analysis uses one single-host common-random-number draw per
(task,~seed) cell. Because the LIBERO/MuJoCo \citep{mujoco2012} rollout is not bit-identical across
hosts, $212$ of the $700$ cells were re-rolled on more than one completion node and
can carry a different success bit per draw, which lets us ask a question the point
estimate alone cannot answer: would the conclusion survive if a different available
draw had defined each cell? We resample, $2{,}000$ times, one available
single-host-complete draw per cell, rebuild the full $700$-pair table, and recompute
the three primary RTR gaps (Fig.~\ref{fig:robustness}). All three contrasts remain
strictly positive in \emph{every} one of the $2{,}000$ redraws: the smallest gap
observed anywhere in the resampling is $+0.003$ for $B{-}C$ and $+0.010$ for $B{-}D$,
and the directional result never reverses. The median resampled gaps run below the
single-draw headline ($B{-}C$ median $0.025$ versus $0.054$; $B{-}D$ median $0.027$
versus $0.050$). The mechanism is specific. When a redraw flips a borderline cell's
weak-arm~(A) outcome to a success, that cell leaves the A-failing conditional pool, and
those borderline cells are exactly the ones targeted escalation is most likely to recover,
so dropping them deflates the measured gap. The robustness
statement is therefore that the \emph{sign and significance} of the selectivity
advantage are invariant to which non-deterministic draw is used, while its
\emph{magnitude} is a conservative function of pool composition; we report the
single-host draw as the headline and this resampling envelope as its sensitivity band.

\FloatBarrier
\section{Discussion}
\label{sec:discussion}

The causal contrasts separate a controller from a difficulty proxy. The result they
license is the one that matters for deployment: the recovered fraction is bought by timing,
not by spending. A difficulty proxy that merely indexes which trajectories are hard would
still beat the weak-only floor, because escalating on hard steps recovers some of them by
brute compute. The budget-matched-blind control (C) and the rate-matched random control (D)
spend that same stronger-policy compute without the probe's timing, and they leave most of
the headroom on the table. So the gap between B and those controls is the part of the
recovery that cannot be explained by extra compute alone. That is what makes the signal a
controller rather than a passive readout, and it is the property an operator actually needs:
the stronger policy earns its cost only on the steps where it changes the outcome.

\begin{figure}[htbp]
  \centering
  \includegraphics[width=0.92\linewidth]{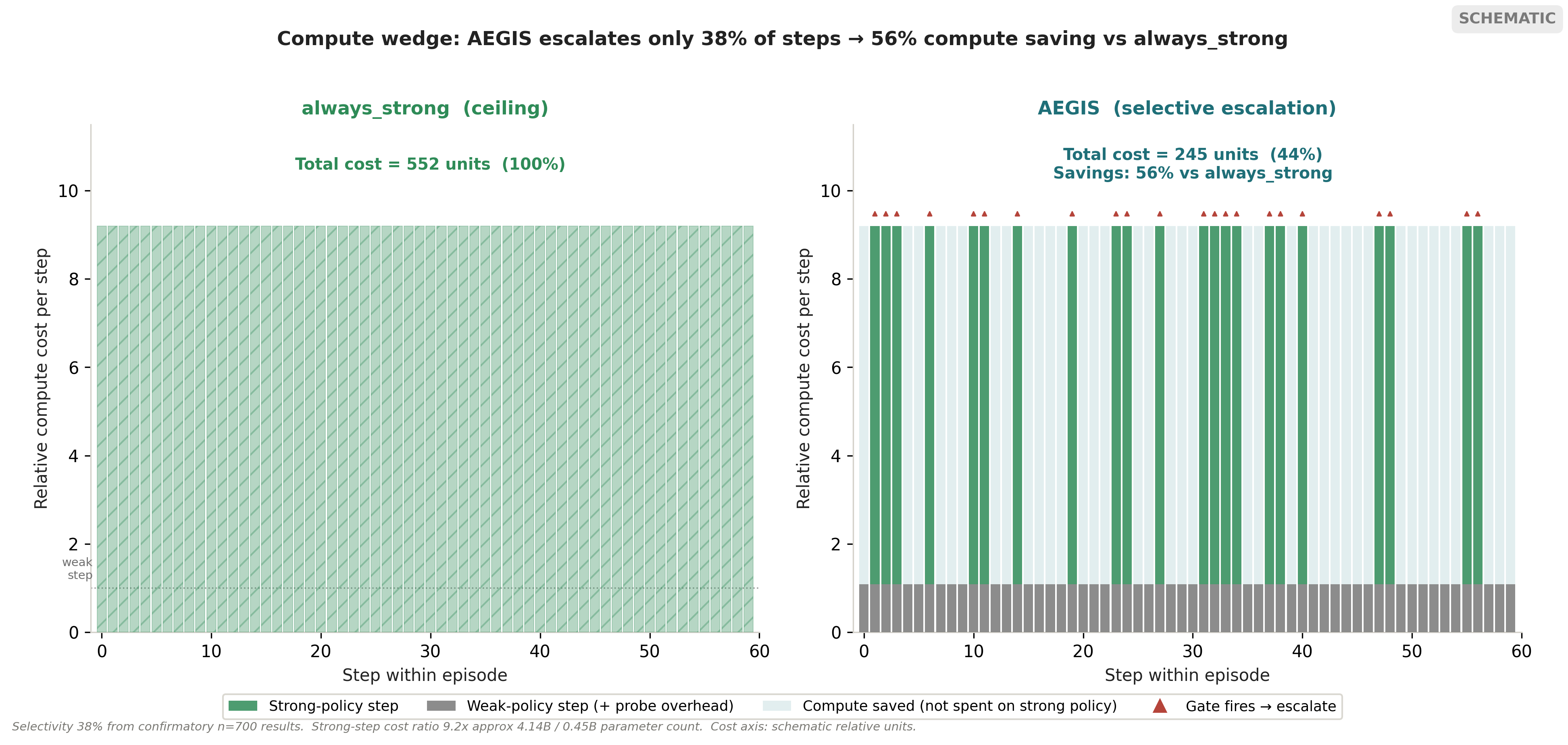}
  \caption{\textbf{Why not just always run the stronger policy.} Per-step compute cost across
  one episode. \emph{Left:} always-strong pays the stronger policy's per-step cost on every
  step. \emph{Right:} AEGIS pays the stronger-policy cost only on the small escalated
  fraction and runs the cheap weak policy (plus probe overhead) elsewhere, so it spends about
  $44\%$ of the always-strong compute, a $56\%$ saving at the confirmatory escalation rate. The
  recovery this compute buys, and the comparison with the budget-matched controls C and D that
  spend the same extra compute without selectivity, is shown on the recovery-versus-compute
  plane of Fig.~\ref{fig:quadrant}. \emph{(Escalated fraction
  ($38\%$ of steps) is from the confirmatory $n{=}700$ run; the compute axis is schematic
  relative units anchored to the $4.14$B/$0.45$B parameter-count ratio.)}}
  \label{fig:cost-wedge}
\end{figure}

The within-stratum test is what closes the remaining loophole. Inside a fixed difficulty
band every arm faces comparably hard trajectories, so a difficulty proxy can no longer
masquerade as a controller. The $n{=}700$ run holds there too: $B>D$ in all three terciles
and $B>C$ in the EASY and MEDIUM bands, with only the HARD-band $B{-}C$ interval touching
zero (Fig.~\ref{fig:hazard}). The pilot and the confirmatory run agree on sign and on the
ordering of the controls, and the attenuation from pilot to full run is the regression
toward a better-estimated value we would expect, not a reversal.
Figure~\ref{fig:cost-wedge} places the recovery against the always-strong ceiling and the
budget-matched controls on a recovery-versus-compute plane.

In a narrow sense this is a form of runtime
\emph{metacognition}: a policy carrying a cheap internal read-out of its own impending
failure and acting on it before the failure compounds. We use the term only as a framing
for the self-monitoring loop. Every claim is grounded in the measured recovered-task-rate,
not in any introspective interpretation of the probe. Framed
against the runtime authorization gap we identify, AEGIS supplies the missing
\emph{authorize-an-escalation} layer: the probe decides, step by step, when the deployed
policy has earned the right to keep driving and when control should pass to a stronger
executor. And it does so under the discipline the intervention-paradox result demands
\citep{interventionparadox2026}: because an accurate predictor can still \emph{reduce}
success when its interventions disrupt trajectories that would have succeeded, we never
argue from predictive AUROC to utility, and we let the budget- and rate-matched controls,
not the ROC curve, carry the causal claim.

\paragraph{Scope discipline and a world-model future extension.}
We deliberately isolate one claim (a frozen-probe early-warning signal plus escalation
to a stronger separate policy, defended with causal controls), and we hold a learned
world model out of scope here. A world model is a natural future extension: a richer
escalation trigger, or a way to choose which stronger policy to escalate to and what to
escalate with \citep{rynnvla2025, farl2026}. Introducing one now would entangle the present
causal contrast with a learned dynamics model, which is exactly what this study is designed
to keep separate.

\section{Limitations}
\label{sec:limitations}

\paragraph{Effect size and remaining headroom.} The headline numbers come from the
confirmatory $n{=}700$ common-random-number factorial ($646$ A-failing), which clears the
pre-registered $\geq 700$-episode floor and the kill criteria (K1--K3). The recovered-task
rate gain over the controls is real but modest ($+0.054$ over budget-matched-blind, $+0.050$
over random-trigger), and the HARD-band $B{-}C$ interval touches zero, so the within-stratum
claim rests on the EASY and MEDIUM bands for the $B>C$ contrast. The exploratory $56$-key
Phase-D pilot ($41$ A-failing) that authorized scaling reported a larger effect; the full
run attenuated it, which we report rather than hide.

\paragraph{Scope: a controlled causal question, with named next tests.} This paper isolates
one question under controlled paired rollouts: does a failure signal choose better
escalation moments than matched-budget controls? It answers that question on LIBERO
\citep{libero2023} with one weak/strong pair
(\texttt{smolvla\_libero}~$\to$~\texttt{pi05\_libero\_finetuned}). Real-hardware transfer,
where perception noise and contact dynamics differ, and broader policy-pair coverage are the
next tests, not assumptions hidden inside the claim. We already take one step on the second
axis: the GR00T~N1.7 generalization arm escalates to a different policy family and recovers
at the same rate as HELM (RTR $0.155$) on the confirmatory run, consistent with the effect
being a property of escalating to a stronger separate policy rather than of this one pair.
A second benchmark suite and a small real-robot demonstration are the highest-value
additions, and they extend the claim rather than underpin it.

\paragraph{Conformal coverage and stratified calibration.} The trigger threshold is
calibrated to a target false-trigger rate, but conformal guarantees are marginal; if
per-stratum calibration is under-powered the realized coverage may drift from the nominal
level within a stratum; the confirmatory $n{=}700$ run reports the within-stratum analysis
at the pre-registered discordant-pair floor, but per-stratum conformal coverage remains
marginal rather than conditional where a stratum was too small to calibrate its own
threshold.

\paragraph{Descriptive, not identified, hazard.} Any hazard-rate or time-to-failure curve
we report is descriptive of when the probe becomes informative; it is not a causal
identification of the failure onset, and we do not use it to make claims beyond the
pre-registered estimand.

\paragraph{Signal fusion is exploratory; overhead does not transfer for free.} The primary
signal is the single hidden-state probe; the surprise/disagreement complements
(\texttt{chunk\_delta}, \texttt{stac\_var}, disagreement) are reported as exploratory
context, not as a tuned fused detector. Finally, the escalation overhead is \emph{measured}
for this weak/strong pair against a profiled baseline; because it scales with the escalated
fraction times the relative per-step cost of the stronger policy, it does not transfer to a
different pair without re-profiling, and we never quote it as a pair-independent constant.

\section{Conclusion}
\label{sec:conclusion}

A deployed policy that is about to fail does not have to keep driving. AEGIS reads a cheap
per-step early-warning probe off a frozen VLA's internals and, on only the steps it flags,
hands control to a stronger separate policy. That is a runtime decision the prior
literature had no layer for: detect-only methods see the failure coming but cannot act,
and recover-in-policy methods act only by asking the same failing policy to try again.

The evidence holds up where it counts. On the confirmatory $n{=}700$ factorial the gain
survives the controls built to break it: against a budget-matched-blind control and a
random-trigger placebo, both of which spend the same extra compute, selective escalation
still adds $+0.054$ and $+0.050$ in conditional recovered-task-rate. The recovery is
bought by timing, not by spending. The early-window failure-prediction AUROC of $0.764$ is
the precondition that makes this cheaper than running the stronger policy on every step,
and we keep it as a precondition rather than a headline, since accurate prediction does not
imply effective prevention. The within-stratum requirement, the always-strong ceiling, the
HELM baseline, and the second-strong-policy generalization arm all came through.

The thesis is one sentence: a robot policy can read its own activations as an
early-warning signal and call a stronger policy before failure compounds, recovering twice
as many failures as matched-budget escalation. A frozen policy can call for backup at the
step where it still matters, and pay for that backup only when it helps. AEGIS makes that
runtime decision, and the controls show the decision is what does the work.

\section*{Data, code, and pre-registration availability}
\addcontentsline{toc}{section}{Data, code, and pre-registration availability}
All artifacts are public. The trained early-warning probe and gate configuration, together
with the frozen pre-registration (analysis plan, primary contrasts, within-stratum floor,
and kill criteria, registered before data collection) and the probe-training,
conformal-calibration, and four-arm common-random-number rollout code, are released as the
model repository \url{https://huggingface.co/kaikaku/aegis}. The per-cell rollout logs
(per-step probe-score traces, policy-source masks, and per-(task, seed, arm) outcomes) that
reproduce Tables~\ref{tab:confirm-rtr}--\ref{tab:harm} and the confirmatory figures are
released as the dataset \url{https://huggingface.co/datasets/kaikaku/aegis-rollouts}. An
interactive demonstration is available at the Space
\url{https://huggingface.co/spaces/kaikaku/aegis-demo}. The analysis scripts recompute the
recovered-task-rate, the exact paired McNemar and binomial tests, the paired-trajectory
bootstrap, and the early-window AUROC directly from the logged traces.

\appendix
\FloatBarrier
\section{Phase-D pilot (exploratory go/no-go)}
\label{app:pilot}
The Phase-D pilot was the pre-committed go/no-go check that authorized the confirmatory
run; we report it here and base no headline claim on it. Its conditional recovered-task
rates are larger than the confirmatory $n{=}700$ values because the $56$-key pilot pool is
small and skews easier, so a single well-timed escalation recovers a larger fraction of it.
The direction and ordering of the arms match the confirmatory run.

\begin{table}[htbp]
  \centering
  \caption{Recovered-task-rate (RTR) and marginal success by arm on the \textbf{Phase-D
  pilot} ($n=56$ CRN keys/arm; $41$ are A-failing and define the conditional estimand;
  horizon $H=10$). RTR is conditional on the weak-only~(A) failing subset. $p$-values are
  Holm-adjusted over the three primary contrasts using the exact McNemar test on paired
  discordant keys. The secondary baseline rows (always-strong, HELM) carry the completed
  $n{=}700$ confirmatory values; the full confirmatory grid is Table~\ref{tab:confirm-rtr}.
  SAFE is detect-only and has no recovery estimand. \emph{Exploratory; not used for the
  headline claim.}}
  \label{tab:main-rtr}
  \begin{tabular}{l c c c c}
    \toprule
    Arm & Marginal success & Conditional RTR & Contrast vs.\ A & Holm $p$ \\
    \midrule
    A. Weak-only            & $26.8\%$ {\scriptsize(15/56)} & $0.000$              & --                  & --            \\
    B. Targeted (method)    & $69.6\%$ {\scriptsize(39/56)} & $\mathbf{0.659}$     & $+0.659$             & $1.5{\times}10^{-8}$ \\
    C. Budget-matched-blind & $28.6\%$ {\scriptsize(16/56)} & $0.146$              & $B{-}C\;{+}0.512$    & $5.7{\times}10^{-6}$ \\
    D. Random-trigger       & $28.6\%$ {\scriptsize(16/56)} & $0.171$              & $B{-}D\;{+}0.488$    & $1.9{\times}10^{-6}$ \\
    \midrule
    Always-strong (ceiling) & $36.0\%$ {\scriptsize($n{=}700$)} & $0.319$ & reference            & --            \\
    HELM-baseline (2ndary)  & $20.5\%$ {\scriptsize($n{=}692$)} & $0.155$ & $B{-}$HELM $-0.055$  & $6.9{\times}10^{-3}$ \\
    SAFE detect-and-halt    & \emph{detect-only} & --        & detection ref        & --            \\
    \bottomrule
  \end{tabular}
\end{table}

\begin{figure}[htbp]
  \centering
  \includegraphics[width=0.95\textwidth]{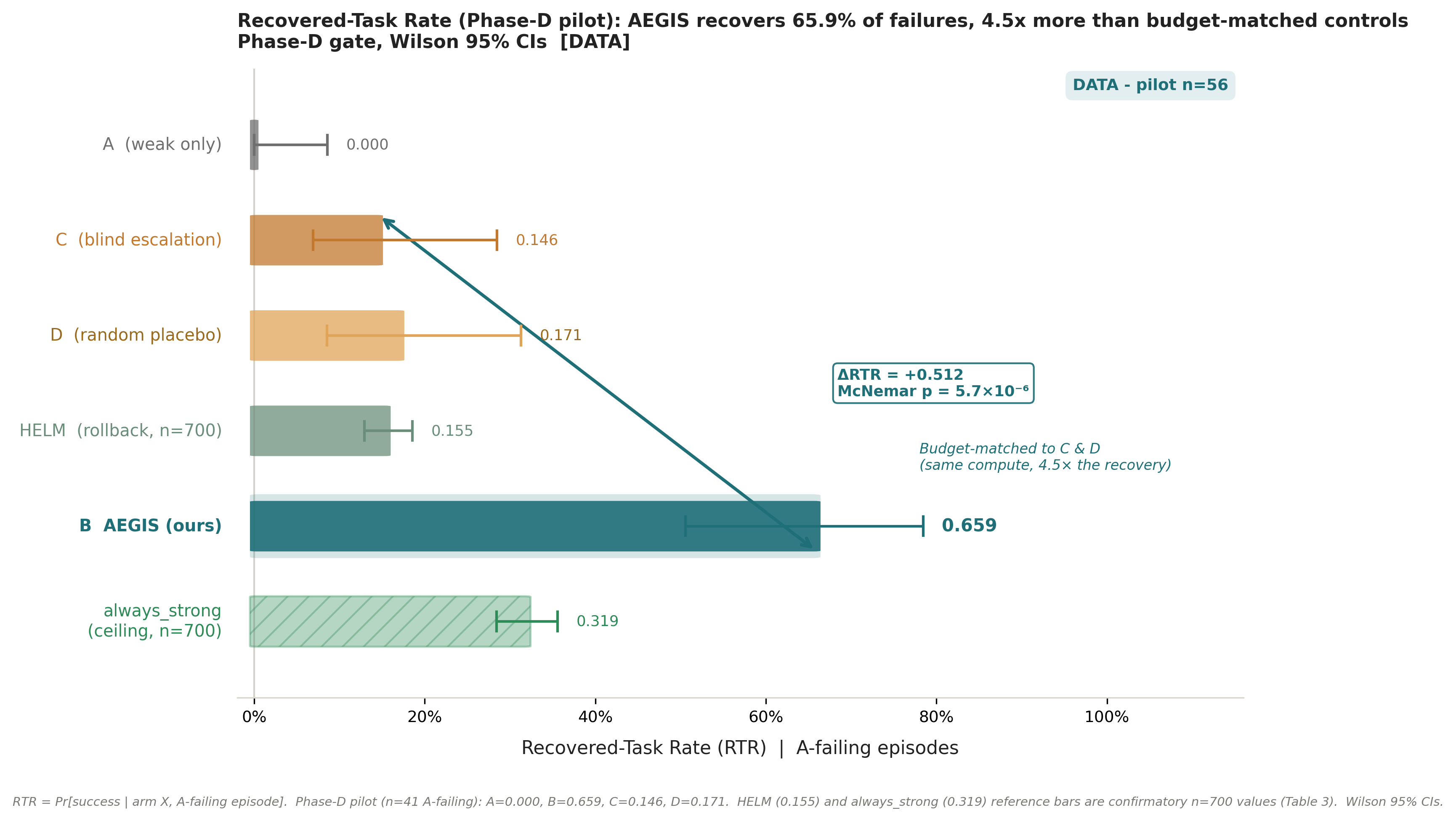}
  \caption{\textbf{Recovered-task rate across arms (Phase-D pilot, exploratory).} RTR
  $=\Pr[\text{success}\mid\text{arm, A-failing episode}]$, with Wilson $95\%$ intervals on
  the $56$-key pilot. Targeted escalation (B) recovers $65.9\%$ of the pilot episodes the
  weak policy alone fails; the compute-matched controls, blind escalation (C, $14.6\%$) and
  the random-trigger placebo (D, $17.1\%$), spend the \emph{same} strong-policy budget yet
  recover far less ($\Delta\mathrm{RTR}_{B-C}=+0.512$, exact McNemar $p=5.7\times10^{-6}$).
  This pilot established directionality only; the headline claim is the confirmatory
  $n{=}700$ result (Fig.~\ref{fig:headline}, Table~\ref{tab:confirm-rtr}).}
  \label{fig:rtr-narrative}
\end{figure}

\FloatBarrier
\bibliographystyle{plainnat}
\bibliography{refs}

\end{document}